\documentclass{article}
\PassOptionsToPackage{table}{xcolor}

\usepackage{iclr2027_conference,times}

\usepackage{amsmath,amsfonts,bm}

\def\eqref#1{equation~\ref{#1}}

\def\1{\bm{1}}

\DeclareMathAlphabet{\mathsfit}{\encodingdefault}{\sfdefault}{m}{sl}
\SetMathAlphabet{\mathsfit}{bold}{\encodingdefault}{\sfdefault}{bx}{n}

\usepackage[utf8]{inputenc} 
\usepackage[T1]{fontenc}    
\usepackage{hyperref}       
\usepackage{url}            
\usepackage{booktabs}       
\usepackage{amsfonts}       
\usepackage{nicefrac}       
\usepackage{tabularx}
\usepackage{array}
\newcolumntype{Y}{>{\raggedright\arraybackslash}X}
\usepackage{microtype}      
\usepackage{graphicx}
\usepackage{subcaption}
\usepackage{xspace}
\newcommand{\BET}{\textsc{Bet}\xspace}

\usepackage{pifont} 
\usepackage{enumitem} 
\usepackage{array}
\usepackage{makecell}
\usepackage{adjustbox}
\usepackage[table]{xcolor}
\usepackage{xspace}
\usepackage[most]{tcolorbox}
\usepackage{amsmath}
\usepackage{enumitem}
\usepackage{amssymb} 
\usepackage{algorithm}
\usepackage{algpseudocode}
\usepackage{booktabs}
\usepackage{multirow}
\usepackage{xcolor}
\usepackage{colortbl}
\usepackage{url}
\usepackage{wrapfig}
\usepackage{graphicx}
\definecolor{routerbg}{HTML}{E8DDF2}    
\definecolor{deerbg}{HTML}{E7D9B6}      
\definecolor{offlinebg}{HTML}{D7E1EC}   
\definecolor{onlinebg}{HTML}{D4E2CD}
\definecolor{betrowbg}{HTML}{B8CCAC}
\definecolor{bbhead}{HTML}{EFEFEC}
\definecolor{dgreen}{HTML}{1B7F3B}
\definecolor{dred}{HTML}{B22222}
\definecolor{dgray}{HTML}{6B6B6B}
\definecolor{dteal}{HTML}{0A5F8A}
\definecolor{accbg}{HTML}{EEF3FA}
\definecolor{tokbg}{HTML}{FBF6EC}
\definecolor{typegray}{HTML}{6B7280}
\definecolor{c1bg}{HTML}{F1E9F7}
\definecolor{c2bg}{HTML}{FBEFDF}
\definecolor{c3bg}{HTML}{E6EEF8}
\definecolor{c4bg}{HTML}{E7F3E6}
\definecolor{betbg}{HTML}{C9E3C6}
\definecolor{refgray}{HTML}{7A7F87}
\definecolor{lightrule}{HTML}{BFC4CA}
\definecolor{dpink}{HTML}{B3123F}
\newcommand{\dsz}{\fontsize{7.5}{8.5}\selectfont}
\definecolor{apphead}{HTML}{E8EEF4}
\definecolor{appalt}{HTML}{F7F9FB}
\usepackage[most]{tcolorbox}
\tcbuselibrary{skins, breakable}
\usepackage{listings}
\usepackage{xcolor}
\usepackage{float}
\definecolor{betboxborder}{HTML}{4E4E4E}
\definecolor{betboxheader}{HTML}{6A6A6A}
\definecolor{betboxbg}{HTML}{F7F7F5}

\lstdefinestyle{betcode}{
  basicstyle=\ttfamily\small,
  breaklines=true,
  breakatwhitespace=false,
  columns=fullflexible,
  keepspaces=true,
  showstringspaces=false,
  frame=none,
  xleftmargin=0pt,
  xrightmargin=0pt,
  aboveskip=0pt,
  belowskip=0pt
}

\newtcolorbox{betexample}[2][]{
  enhanced,
  breakable,
  colback=betboxbg,
  colframe=betboxborder,
  boxrule=0.9pt,
  arc=3pt,
  outer arc=3pt,
  left=8pt,
  right=8pt,
  top=7pt,
  bottom=7pt,
  title=#2,
  colbacktitle=betboxheader,
  coltitle=white,
  fonttitle=\bfseries,
  attach boxed title to top left={xshift=0pt,yshift=0pt},
  boxed title style={
    size=small,
    colframe=betboxborder,
    boxrule=0.9pt,
    arc=3pt,
    outer arc=3pt
  },
  #1
}

\newcommand{\yes}{\textcolor{dgreen}{\ding{51}}}
\newcommand{\no}{\textcolor{dgray}{\ding{55}}}
\newcommand{\legendbox}[1]{%
  \fcolorbox{black!30}{#1}{\rule{0pt}{0.55em}\rule{0.55em}{0pt}}%
}

\title{\textit{Nice Fold} or \textit{Hero Call}: Learning Budget-Efficient Thinking under Policy-Dependent Solvability}

\author{Zhaomeng Zhou$^{\spadesuit}$, Lan Zhang$^{\spadesuit\ast}$, Junyang Wang$^{\spadesuit}$, Mu Yuan$^{\heartsuit}$, Songlin Liu$^{\clubsuit}$, \\
\bf Tingzhao Li$^{\clubsuit}$, Yiqing Hu$^{\clubsuit\ast}$, Yumeng Zhao$^{\spadesuit}$ \\
{\normalfont\normalsize $^{\spadesuit}$University of Science and Technology of China \quad $^{\heartsuit}$The Chinese University of Hong Kong} \\
{\normalfont\normalsize $^{\clubsuit}$ChangXin Memory Technologies, Inc.} \\
{\normalfont\normalsize \texttt{zhouzhm@mail.ustc.edu.cn}, \texttt{zhanglan@ustc.edu.cn}, \texttt{muyuan@cuhk.edu.hk}}}

\iclrfinalcopy
\begin{document}

\maketitle
\lhead{}\renewcommand{\headrulewidth}{0pt}
\renewcommand{\thefootnote}{\fnsymbol{footnote}}
\footnotetext[1]{Corresponding authors.}
\renewcommand{\thefootnote}{\arabic{footnote}}

\begin{abstract}
Large reasoning models (LRMs) improve problem solving through extended reasoning, but often misallocate test-time compute. Existing efficiency methods reduce cost by compressing reasoning traces or conditioning budget on perceived difficulty, yet read the resulting pass rate as a difficulty score, leaving the zero-return regime unmodeled. As a result, they overspend on queries beyond the model's capability while compressing hard-but-solvable ones that need deeper reasoning. In this work, we formulate adaptive reasoning as a computational investment under uncertainty, where budget follows the expected return of reasoning rather than perceived difficulty. To instantiate this principle, we propose \textbf{B}udget-\textbf{E}fficient \textbf{T}hinking (\BET), a two-stage framework that combines behavioral cold-start with GRPO under an investment-cost-aware reward. By aligning solve-or-fold decisions with rollout-derived solvability, \BET learns three behaviors: (1) \textit{short solve}, answering easy queries concisely; (2) \textit{nice fold}, abstaining early when continued reasoning has near-zero expected return; and (3) \textit{hero call}, preserving sufficient compute for hard-but-solvable queries. Across seven benchmarks and three base models, \BET reduces reasoning tokens by \textbf{54\%} while improving up to \textbf{3.2\%} accuracy, and transfers zero-shot to scientific QA and logical reasoning with comparable efficiency gains. Code is available at \url{https://github.com/houqiii/BET}.
\end{abstract}

\section{Introduction}

LRMs such as OpenAI-o1~\citep{elkishky2024openaios} and DeepSeek-R1~\citep{deepseekai2025deepseekr1ir} have advanced complex problem solving through extended reasoning with multi-step verification~\citep{Wei2022ChainOT,zhang2025aso}. As these models become central to real-world agentic systems~\citep{Watanabe2025OnTU,Wang2026OpenClawRLTA,Zhou2025IoTBrainGL}, their inference-time cost is becoming unsustainable~\citep{Sui2025StopOA,Feng2025EfficientRM}. Part of this cost is the genuine price of solving hard problems, and the rest stems from overthinking on easy queries and futile exploration beyond the model's competence~\citep{Chen2024DoNT,liu2025diffadaptdr,Wu2025WhenMI}. The central challenge is to \textit{allocate test-time compute according to its expected return without sacrificing problem-solving capability.}

Recent work addresses this burden in two main ways: \textit{uniform reasoning compression} reduces generation cost through concise-chain distillation~\citep{Chen2024DoNT}, length shaping~\citep{Shi2025EfficientRF,Tu2025LearningWT}, or target-chain supervision~\citep{Chen2025VeriThinkerLT,Arora2025TrainingLM}, while \textit{difficulty-conditioned budget control} acts more finely by injecting early-exit signals during decoding~\citep{Yang2025DynamicEE} or learning query-aware depth policies~\citep{Zhu2025TheLA,Huang2025AdaCtrlTA,Chen2025AwareFT}. These methods reduce token usage by compressing reasoning globally or scaling its depth to perceived difficulty, turning adaptive reasoning into a problem of how much compute to spend on each query.

\begin{figure}[t]

    \centering
    \includegraphics[width=\linewidth]{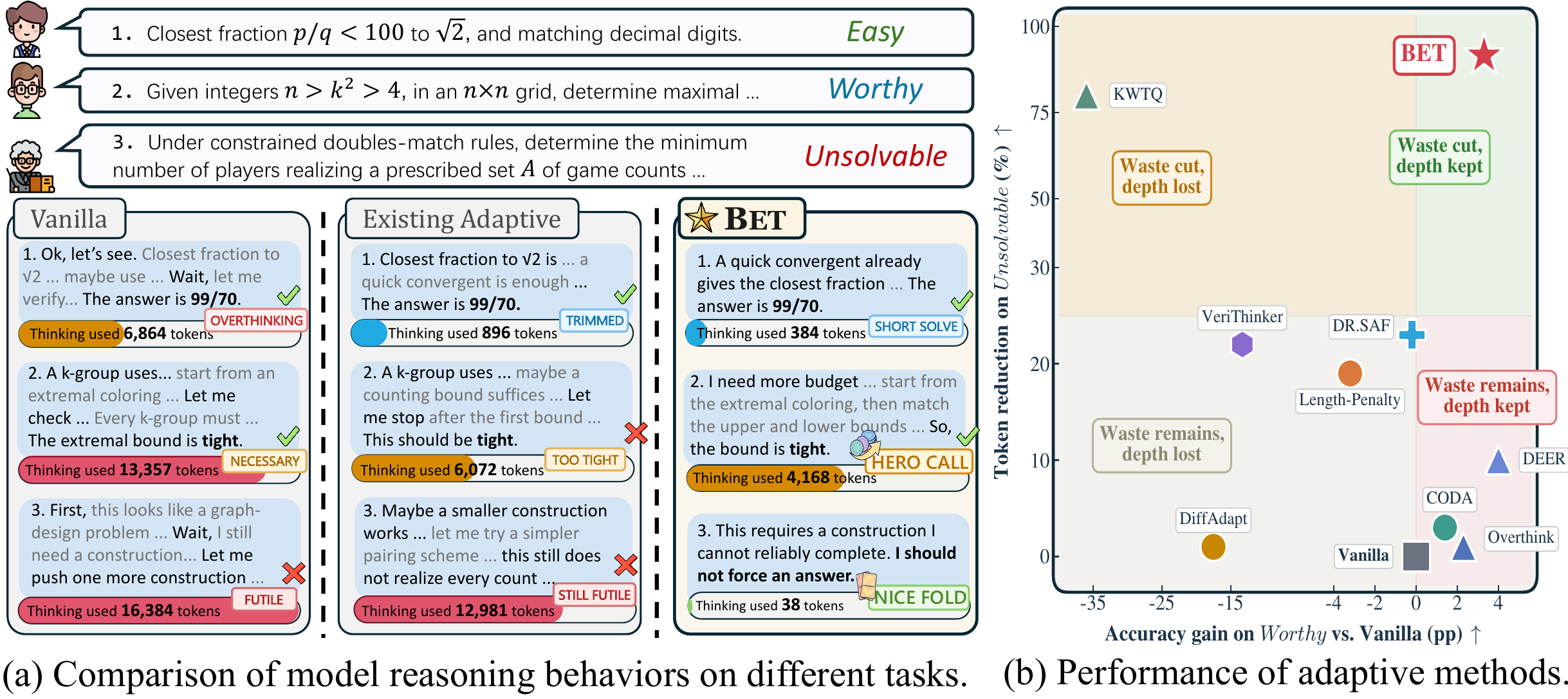}
    \vspace{-0.23in}
\caption{\textbf{Reasoning behavior and accuracy-efficiency trade-offs on Omni-Math.} \textbf{(a)} \textit{Easy}, \textit{worthy}, and \textit{unsolvable} correspond to 16/16, 1--15/16, and 0/16 correct vanilla rollouts. Vanilla LRMs overthink on easy queries and over-allocate on unsolvable ones, while prior adaptive methods often curtail worthy reasoning prematurely. \textbf{(b)} \BET lies on the Pareto frontier, preserving \textit{worthy} reasoning while reducing waste on \textit{unsolvable} queries.}
    \vspace{-0.5cm}
    \label{fig:fig1}
\end{figure}

In realistic workloads, however, difficulty is not solvability. Problems with similar nominal difficulty can have opposite return profiles under the current policy, with one yielding to deeper reasoning and another remaining unsolved even at full budget~\citep{snell2024scaling}. Existing methods often treat hard queries as longer versions of easy ones, which reflects a conflation of solvability with difficulty. For a given policy, hard queries include both hard-but-solvable cases, where deeper reasoning can change the answer, and effectively unsolvable cases, where success remains negligible even with large budgets~\citep{brown2024large}. As Fig.~\ref{fig:fig1}(a) illustrates, this missing signal leads to two errors. \textbf{(1)} Models spend near-maximal budgets on unsolvable queries with near-zero empirical return. \textbf{(2)} Under global or difficulty-conditioned compression, they also shorten hard-but-solvable queries, losing correctness that sufficient reasoning would have preserved.

These observations motivate a shift from difficulty-conditioned compression to solvability-aware budget investment. We instantiate this view in \textbf{B}udget-\textbf{E}fficient \textbf{T}hinking (\BET), a two-stage framework that allocates test-time compute by policy-dependent solvability. \BET estimates whether further investment is likely to pay off under the current policy and learns three target behaviors: \textbf{(1)}~\textit{short solve}, answering easy queries concisely; \textbf{(2)}~\textit{nice fold}, abstaining early when continued reasoning has near-zero expected return; and \textbf{(3)}~\textit{hero call}, investing sufficient compute in hard-but-solvable queries where deeper reasoning can still change the answer.

Realizing this policy is non-trivial because solvability and efficient cost are posterior signals, while budget investment must be committed before reasoning begins. \BET bridges this gap by converting current-policy rollouts into a group profile and optimizing a composite investment-cost-aware reward. The reward gates solve-or-abstain value by empirical solvability, applies query-adaptive efficiency shaping to shorten redundancy without penalizing necessary exploration, and finally calibrates pre-reasoning solvability and budget estimates to the rollout-derived profile. The resulting policy plans its token budget before reasoning, folds selectively on queries it cannot solve, and preserves depth on hard-but-solvable ones. Together, these behaviors improve the accuracy-efficiency trade-off and place \BET on the Pareto frontier in Fig.~\ref{fig:fig1}(b). We summarize our contributions as follows:

\begin{itemize}[leftmargin=1.5em]
    \item[\ding{182}] \textbf{Solvability-Aware Compute Allocation.}
    We recast adaptive reasoning as compute investment under uncertainty, giving each
    query a budget that follows the return it can still yield under the current policy,
    estimated online from that policy's own rollouts.
 
    \item[\ding{183}] \textbf{Investment-Aware Training.}
    We propose \BET, which pairs behavioral cold-start with investment-cost-aware
    GRPO. An asymmetric solvability gate, efficiency shaping, and budget calibration
    jointly induce \textit{short solve}, \textit{nice fold}, and \textit{hero call} reasoning behavior.
 
    \item[\ding{184}] \textbf{A Better Accuracy-Efficiency Frontier.}
    Across seven benchmarks and three base models, \BET is
    \textbf{(I)}~\textit{\textbf{cost-effective}}, reducing tokens by 54\%
    with accuracy gains reaching 3.2\%;
    \textbf{(II)}~\textit{\textbf{waste-reducing}}, cutting unsolvable-query tokens
    by 92\% while preserving solvable depth; and
    \textbf{(III)}~\textit{\textbf{generalizable}}, transferring to scientific QA
    and logical reasoning with 56\% token savings and consistent gains.
\end{itemize}

\section{Related Work}
\label{sec:related}

\paragraph{Uniform reasoning compression.}
Overthinking in LRMs has motivated methods that reshape reasoning toward concise solutions. QFFT~\citep{Liu2025QFFTQF} preserves short-chain instincts by removing SFT questions, CoT-Valve~\citep{Ma2025CoTValveLC} interpolates between long- and short-CoT models, and VeriThinker~\citep{Chen2025VeriThinkerLT} suppresses redundant self-reflection through auxiliary verification. Others identify overthinking transitions or optimize preferences toward concise paths~\citep{Zhao2025LetLB,An2025DontTL}, while length-based RL uniformly penalizes generation cost~\citep{Arora2025TrainingLM,Luo2025O1PrunerLF,Aggarwal2025L1CH} and inference-time variants steer activations~\citep{Huang2025MitigatingOI} or compress prompts~\citep{Xu2025ChainOD,Han2024TokenBudgetAwareLR}. These methods set the budget by a single global pressure that ignores the individual query. \textit{Such uniform pressure shortens reasoning, but it applies the same discount to redundant verification and to exploration that would still have changed the answer.}

\paragraph{Difficulty-conditioned budget control.}
A second family of methods makes the budget query-dependent. Preflight controllers decide in advance: ThinkSwitcher~\citep{Liang2025ThinkSwitcherWT} and DiffAdapt~\citep{liu2025diffadaptdr} route queries across modes, and others preset token budgets~\citep{Zhao2025SABERSA,Wen2025BudgetThinkerEB}. In-generation controllers act during decoding: DEER~\citep{Yang2025DynamicEE} exits on confidence stabilization, while KWTQ~\citep{davidov2026knowingquitprincipledframework} abstains below an auxiliary value threshold. Training-based controllers write difficulty into the objective: DR.SAF~\citep{Chen2025AwareFT} assigns incentives from grouped pass rates, AdaCtrl~\citep{Huang2025AdaCtrlTA} thresholds them into tagged modes, and CODA~\citep{wu2026codadifficultyawarecomputeallocation} gates the reward on online difficulty. Effort still scales along a single difficulty axis, where hard-but-solvable and effectively unsolvable queries share the low end. \textit{\BET instead gives the policy an explicit abstain action, gates it asymmetrically on rollout evidence, and calibrates it jointly with the token budget, so effort follows attainable return.}

\section{Method} 
\label{sec:method}

\subsection{Problem Setup}
\label{sec:compute_allocation}

Consider a reasoning policy $\pi_\theta$ that, given a query $x \in \mathcal{X}$, generates a response $y \sim \pi_\theta(\cdot \mid x)$ at a token cost $c(y) \in \mathbb{R}_+$. Queries differ in how amenable they are to the model, which we capture through the \textit{policy-dependent solvability}:
\begin{equation}
\label{eq:solvability}
    s_\pi(x) \;=\; \Pr\nolimits_{y \sim \pi(\cdot \mid x)}\!\bigl[\,\textsc{Correct}(y)\,\bigr],
\end{equation}
which is the probability that the current policy produces a correct solution for $x$. This quantity is policy-dependent rather than query-intrinsic. It evolves during training and may shift from near zero to positive as the model improves. Accordingly, we call a query \textit{unsolvable} when its empirical success rate under the current policy is negligible, indicating a capability mismatch rather than the absence of a valid solution. We formalize the threshold in \S~\ref{sec:reward} and justify it in Appendix~\ref{app:fold_discussion}.

Under this view, each query becomes a \textit{compute investment decision} in which the policy balances the expected value of a correct solution against failed exploration and token cost. We formalize this trade-off as maximizing the expected net return over the query distribution $\mathcal{D}$,
\begin{equation}
\label{eq:objective}
    \max_{\pi} \;\; \mathbb{E}_{x \sim \mathcal{D}}\!\Big[\,
        s_\pi(x) \cdot r^{+}
        \;-\; \bigl(1 - s_\pi(x)\bigr) \cdot \phi\!\bigl(\bar{c}_\pi(x)\bigr)
        \;-\; \alpha_{\mathrm{cost}} \cdot \bar{c}_\pi(x)
    \,\Big],
\end{equation}
\noindent where $r^{+}$ is the reward for a correct solution, $\phi(\cdot)$ is a cost-dependent penalty for failed attempts, $\bar{c}_\pi(x)=\mathbb{E}_{y \sim \pi}[c(y)]$ is the expected token cost, and $\alpha_{\mathrm{cost}}>0$ controls the global cost pressure.

\paragraph{Budget investment depends on expected reasoning return.}
When $s_\pi(x)\approx0$, the expected value of solving vanishes and failed attempts dominate, so the policy should stop early. When $s_\pi(x)>0$, additional budget is justified only while reasoning can still improve the answer. The hard case lies between these extremes, where possible gains must be weighed against token cost. Since both $s_\pi(x)$ and marginal return are unknown before generation, Eq.~\ref{eq:objective} requires the policy to estimate its own competence and budget need before reasoning begins.

\subsection{When to Invest: Policy-Dependent Solvability}
\label{sec:dynamic_estimation}

\begin{tcolorbox}[colback=cyan!6, colframe=cyan!40!gray, boxrule=0.4pt, arc=2pt, left=6pt, right=6pt, top=2pt, bottom=2pt]
\textbf{\textit{Key Observation 1.}}\;
Solvability is policy-dependent. The same query may be solvable under one policy yet beyond reach under another, since success depends jointly on the query and on current model capability. Thus, fixed difficulty labels cannot provide stable allocation targets.
\end{tcolorbox}

\vspace{2pt}
Solvability must therefore be estimated from policy behavior. External difficulty labels, whether human ratings or competition hierarchies~\citep{Gao2024OmniMATHAU,lightman2023lets,balunovic_srimatharena_2025}, do not track a model-specific success boundary~\citep{Chen2025AwareFT,liu2025diffadaptdr}, while recent adaptive methods read grouped rollout pass rates as query-hardness proxies~\citep{Chen2025AwareFT,Huang2025AdaCtrlTA,wu2026codadifficultyawarecomputeallocation}. \BET instead reads this statistic as current-policy solvability, measuring whether continued reasoning is likely to yield return under the policy being optimized, and recomputes it from $\pi_\theta$ throughout training. Given $K$ independent rollouts $\{y_1,\dots,y_K\}$ on query $x$, we derive two statistics that characterize compute-worthiness.

\paragraph{Group success rate.}
We estimate solvability with its Monte Carlo counterpart,
\begin{equation}
\label{eq:group_success}
    \hat{s}(x) \;=\; \frac{1}{K}\sum_{k=1}^{K} \mathbf{1}\!\bigl[\textsc{Correct}(y_k)\bigr].
\end{equation}
Rollouts come from the current $\pi_\theta$, so $\hat{s}(x)$ is non-stationary. Queries may move from $\hat{s}(x)\approx0$ to $\hat{s}(x)>0$ as the model improves, leaving the zero-return regime without relabeling. The label is one-sided. It requires all $K$ rollouts to fail, an event of probability $(1{-}s)^{K}$ that decays geometrically in $K$. At $K{=}16$ this probability falls below $5\%$ once $s\geq0.17$, so a query wrongly assigned there has low true solvability and, by Eq.~\ref{eq:objective}, low expected return. Empirically, only $0.43\%$ of zero labels flip when $K$ grows from $16$ to $24$ (Fig.~\ref{fig:fig5}(c)), as analyzed in Appendix~\ref{app:fold_discussion}.

\paragraph{Efficient solution cost.}
For queries with $\hat{s}(x)>0$, we estimate a lower-envelope cost of successful reasoning under the current policy. Let $\mathcal{C}(x)=\{y_k:\textsc{Correct}(y_k)\}$ denote correct rollouts sorted by token length. We define the \textit{efficient solution cost} $\hat{c}^{*}(x)$ as the mean length of the shortest $m(x)=\max\{1,\lceil p|\mathcal{C}(x)|\rceil\}$ trajectories in $\mathcal{C}(x)$, with $p=30\%$. This lower-envelope estimate filters long correct trajectories and marks the cost beyond which further reasoning has diminishing return.

\subsection{How to Invest: An Asymmetric Compute-Aware Reward}
\label{sec:reward}

\begin{tcolorbox}[colback=cyan!6, colframe=cyan!40!gray, boxrule=0.4pt, arc=2pt, left=6pt, right=6pt, top=2pt, bottom=2pt]
\textbf{\textit{Key Observation 2.}}\;
Within the solvable range, additional reasoning eventually becomes low-return, creating a natural budget ceiling between productive and redundant exploration.
\end{tcolorbox}

\vspace{2pt}

\begin{wrapfigure}{r}{0.50\textwidth}
\vspace{-0.45cm}
\centering
\includegraphics[width=\linewidth]{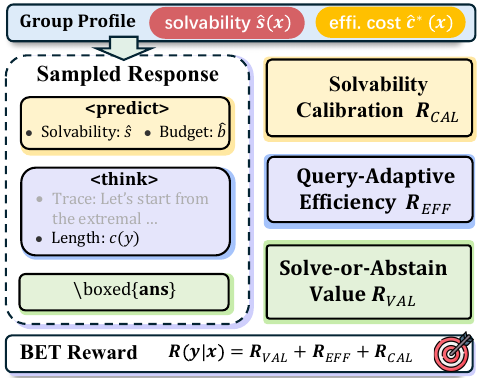}
\vspace{-0.40cm}
\caption{\textbf{Composite reward structure in \BET.} A sampled response splits into \texttt{<predict>}, \texttt{<think>}, and answer components, shaped by $R_{\textsc{cal}}$, $R_{\textsc{eff}}$, and $R_{\textsc{val}}$ under the group profile.}
\label{fig:bet_reward_structure}
\vspace{-0.75cm}
\end{wrapfigure}

The two observations together split the query space into a zero-return zone, where abstention dominates, and a diminishing-return zone, where reasoning is useful only up to a finite cost ceiling. An efficient policy should solve easy queries concisely (\textit{short solve}), preserve depth for hard-but-solvable ones (\textit{hero call}), and abstain on unsolvable ones (\textit{nice fold}), so the reward must encode which action to take before reasoning begins as well as correctness and cost.

As shown in Fig.~\ref{fig:bet_reward_structure}, we carry Eq.~\ref{eq:objective} into training as a per-trajectory reward $\mathcal{R}(y \mid x)=R_{\textsc{val}}+R_{\textsc{eff}}+R_{\textsc{cal}}$. $R_{\textsc{val}}$ takes up its solution value and failed-attempt penalty, $R_{\textsc{eff}}$ replaces its global cost term with query-adaptive efficiency shaping, and $R_{\textsc{cal}}$ supplies the estimates it presupposes. All terms use the group profile $(\hat{s}(x),\hat{c}^{*}(x))$, so the reward evolves with competence rather than static annotations.

\paragraph{Solution value with abstention gate.}
We augment the standard correctness reward with an explicit \textit{nice fold} action. A correct solution earns reward $+1$, and an incorrect attempt incurs the cost-proportional penalty $-\phi(c(y))$. Abstention is gated asymmetrically by solvability:
\begin{equation}
\label{eq:rval}
R_{\textsc{val}}(y \mid x)=
\begin{cases}
+1, & y \text{ is correct},\\
-\phi(c(y)), & y \text{ is incorrect},\\
R_{\textsc{abstain}}(x), & y \text{ abstains},
\end{cases}
\qquad
R_{\textsc{abstain}}(x) =
\begin{cases}
+\delta, & \hat{s}(x) < \epsilon_{\mathrm{abs}},\\
-\lambda, & \hat{s}(x) \ge \epsilon_{\mathrm{abs}}.
\end{cases}
\end{equation}
Here $\delta>0$ rewards identifying an unsolvable query, $\lambda\gg\delta$ penalizes premature surrender, $\phi(\cdot)$ is a monotone cost penalty on failed attempts, and $\epsilon_{\mathrm{abs}} = 1/K$ makes abstention rewarding only when all $K$ rollouts fail (\S~\ref{sec:dynamic_estimation}). We instantiate $\phi(c)=\alpha_{\mathrm{fail}}\cdot c/L_{\max}$, with $L_{\max}$ the maximum completion length and $\alpha_{\mathrm{fail}}$ the failed-attempt cost weight.

This asymmetry keeps fold from becoming an absorbing shortcut. With all rollouts failed, abstention yields $+\delta$. Beyond leaving each rollout free to reason in depth, we keep one attempting whenever all $K$ would abstain, so a success stays observable (Appendix~\ref{app:algorithm}). Once any rollout succeeds, $\hat{s}(x) \ge \epsilon_{\mathrm{abs}}$ and abstention receives $-\lambda$, while correct solutions receive $+1$. Since $\lambda\gg\delta$, a single success flips abstention from reward to penalty, preserving incentives for \textit{hero call} as verified in \S~\ref{sec:emergence}.

\paragraph{Query-adaptive efficiency shaping.}
Efficiency pressure should shorten redundancy without penalizing necessary exploration. We therefore condition length shaping on correctness and efficient solution cost, rewarding only correct solutions shorter than the current cost estimate,
\begin{equation}
\label{eq:reff}
    R_{\textsc{eff}}(y \mid x) = \begin{cases}
        \displaystyle \beta \cdot \max\!\Bigl(0,\; 1 - \frac{c(y)}{\hat{c}^{*}(x)}\Bigr) & y \text{ is correct} \;\wedge\; \hat{s}(x) > \tau, \\[6pt]
        0 & \text{otherwise},
    \end{cases}
\end{equation}
where $\hat{c}^{*}(x)$ is the efficient solution cost from \S~\ref{sec:dynamic_estimation}, $\tau$ is a confidence threshold, and $\beta$ controls bonus strength. For easy queries with small $\hat{c}^{*}(x)$, the bonus is positive only for short correct solutions, encouraging concision. For hard-but-solvable queries whose $\hat{c}^{*}(x)$ approaches the budget, the bonus vanishes and deep reasoning remains unpenalized. $R_{\textsc{eff}}$ activates only for correct solutions with $\hat{s}(x)>\tau$, since failed attempts are already cost-shaped by $R_{\textsc{val}}$. This avoids double penalties, while $\tau>\epsilon_{\mathrm{abs}}$ prevents unstable low-solvability estimates from setting the efficiency target.

\paragraph{Solvability calibration.}
Eq.~\ref{eq:objective} requires allocation before true solvability and efficient cost are known, so the model must commit to a structured pre-reasoning estimate with predicted solvability $\hat{s}$ and budget $\hat{b}$ in $[0,1]$. $R_{\textsc{cal}}$ calibrates these estimates against the group profile $(\hat{s}(x),\hat{c}^{*}(x))$,
\begin{equation}
\label{eq:rcal}
    R_{\textsc{cal}}(y \mid x) = \begin{cases}
        \displaystyle -\gamma_s \!\cdot\! \bigl|\hat{s} - \hat{s}(x)\bigr| \;-\; \gamma_b \!\cdot\! \ell\!\bigl(\hat{b},\, b^{*}(x)\bigr)
            & \hat{s}(x) \ge \epsilon_{\mathrm{abs}}, \\[6pt]
        \displaystyle -\gamma_s' \!\cdot\! \hat{s} \;-\; \gamma_b' \!\cdot\! \hat{b}
            & \hat{s}(x) < \epsilon_{\mathrm{abs}},
    \end{cases}
\end{equation}
where $b^{*}(x)=\hat{c}^{*}(x)/L_{\max}$ is the normalized budget target, $\ell(\hat{b}, b^{*})$ penalizes budget underestimation more than overestimation, and $\gamma_s' > \gamma_s$ strengthens zero-return calibration. When all rollouts fail, this term pushes predicted solvability toward zero and suppresses declared budget. $R_{\textsc{cal}}$ shapes capability representation and supplies the planning signal for compute allocation, leaving the solve-or-abstain choice to $R_{\textsc{val}}$. Definitions of $b^{*}(x)$ and $\ell(\hat{b}, b^{*})$ are in Appendix~\ref{app:aux_symbols}.

\subsection{Two-Stage Training}
\label{sec:training}

\paragraph{Stage 1: Behavioral cold-start.}
Directly optimizing the composite reward from a vanilla model is impractical because the policy lacks the behaviors reward differentiation needs: structured solvability estimation, budget-aware generation, and rational abstention. We address this with SFT on curated demonstrations that split correctly answered queries into \textit{short solve} and \textit{hero call} by offline cost profiling, and add \textit{nice fold} for the queries a single base-policy sample fails, where the model emits a designated \texttt{<Unsolvable>} token. This stage is purely behavioral. It exposes the policy to the full action space, including formatting, estimation syntax, and abstention, while the statistics that drive the reward are estimated from $K$ rollouts online in Stage~2 (Appendix~\ref{app:sft}).

\paragraph{Stage 2: RL with investment-cost-aware reward.}
We instantiate the reward of \S~\ref{sec:reward} within GRPO~\citep{Shao2024DeepSeekMathPT}. At each step the current policy generates $K=16$ rollouts per query, $\hat{s}(x)$ and $\hat{c}^{*}(x)$ are computed on the fly, the composite reward is evaluated per trajectory, and the policy is updated with clipped surrogate optimization and group-relative advantages, so the reward tracks the evolving capability boundary. Algorithm~\ref{alg:bet_stage2} gives the procedure.

\subsection{Emergence of the Three Behaviors}
\label{sec:emergence}

We verify that the three target behaviors emerge after training using a fixed pre-\BET diagnostic. With the frozen base Qwen3-4B policy (\textit{vanilla}) on Omni-Math~\citep{Gao2024OmniMATHAU}, we run $K{=}16$ rollouts per query, matching the training-time profiling granularity, and partition queries into \textit{easy} (16/16), \textit{worthy} (1--15/16), and \textit{unsolvable} (0/16) regimes. This split marks where reasoning is already reliable, potentially answer-changing, or futile, and tests whether \BET reallocates compute accordingly.

\begin{figure*}[t]
    \centering
    \includegraphics[width=\textwidth]{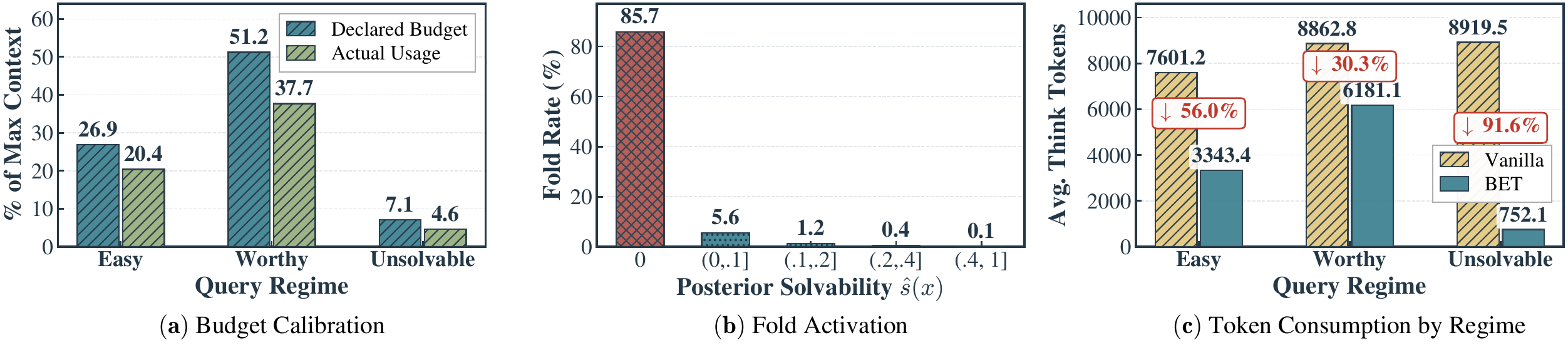}
    \vspace{-0.18in}
    \caption{\textbf{Behavioral diagnostics of \BET on Omni-Math.} \textbf{(a)} Declared and realized budget by query regime, as percentages of the maximum context length. \textbf{(b)} Fold rate versus vanilla posterior solvability $\hat{s}_0(x)$, concentrated at $\hat{s}_0(x){=}0$. \textbf{(c)} Average think tokens for vanilla and \BET by regime.}
    \vspace{-0.4cm}
    \label{fig:fig2}
\end{figure*}

\paragraph{\ding{182} \BET plans and differentiates compute by query solvability.}
The vanilla policy spends nearly the same budget on \textit{worthy} and \textit{unsolvable} queries, treating both as similarly demanding. In contrast, \BET separates them in Fig.~\ref{fig:fig2}(c), eliminating 91.6\% of \textit{unsolvable} expenditure through learned abstention while retaining 70\% of the \textit{worthy} budget for productive reasoning. It also shortens \textit{easy} query generation through \textit{short solve}. Declared budgets track usage (Fig.~\ref{fig:fig2}(a)), so the model commits to a budget before it starts generating.

\paragraph{\ding{183} \BET concentrates its folds at the zero-return boundary.}
Folding is sharply localized where Observation~1 predicts little return. Fig.~\ref{fig:fig2}(b) shows 85.7\% of rollouts abstaining at $\hat{s}_0(x){=}0$, with the rate falling almost to zero once $\hat{s}_0(x)>0.2$. This step pattern localizes fold at the empirical capability boundary. Fold and \textit{hero call} also coexist without interference. Under the same \textit{worthy} partition, \BET obtains the largest net gain in correct answers among all compared methods (Fig.~\ref{fig:fig4}(b), \S~\ref{sec:regime}), so abstention reclaims budget from futile queries while leaving productive reasoning intact.

\section{Experiments}
\label{sec:experiments}

\subsection{Experimental Setup}
\label{sec:setup}

\paragraph{Training details.}
We use Qwen3-4B-Thinking-2507~\citep{qwen3_4b_thinking_2507_hf}, DeepSeek-R1-Distill-Qwen-7B~\citep{deepseek_r1_distill_qwen_7b_hf}, and DeepSeek-R1-Distill-Qwen-14B~\citep{deepseek_r1_distill_qwen_14b_hf}. Both stages train on a stratified 20K subset of DeepMath-103K~\citep{He2025DeepMath103KAL}, sampled uniformly across difficulty labels and decontaminated against all evaluation benchmarks. Stage~1 uses LoRA~\citep{Hu2021LoRALA} SFT to induce structured solvability estimation and the three target behaviors. Stage~2 applies GRPO~\citep{Shao2024DeepSeekMathPT} with TRL~\citep{vonwerra2020trl}, using $K{=}16$ rollouts, temperature $0.8$, and maximum length $16{,}384$, with $\beta$ and $p$ kept at their defaults of $0.30$ and $30\%$ throughout. Length-Penalty, CODA, and DR.SAF use the same group size and rollout protocol for fair comparison. Details are in Appendix~\ref{app:hyperparams}. 
 
\paragraph{Baselines.}
We compare \BET against eight baselines, all re-implemented with the same backbones and decoding configuration: \textbf{(i)}~\textit{preflight routing} (DiffAdapt~\citep{liu2025diffadaptdr}), \textbf{(ii)}~\textit{in-generation intervention} (DEER~\citep{Yang2025DynamicEE}, KWTQ~\citep{davidov2026knowingquitprincipledframework}), \textbf{(iii)}~\textit{offline adaptation} (VeriThinker~\citep{Chen2025VeriThinkerLT}, Overthink~\citep{Chen2024DoNT}), and \textbf{(iv)}~\textit{online post-training} (Length-Penalty~\citep{Arora2025TrainingLM}, CODA~\citep{wu2026codadifficultyawarecomputeallocation}, DR.SAF~\citep{Chen2025AwareFT}).

\subsection{Evaluation}
\label{sec:eval}

\paragraph{Benchmarks.}
We evaluate on seven reasoning benchmarks in two groups. For \textit{in-domain} mathematical reasoning, we use Omni-Math~\citep{Gao2024OmniMATHAU}, MATH500~\citep{lightman2023lets}, AMC-23~\citep{mathai_amc23_2025}, and AIME-25~\citep{balunovic_srimatharena_2025}, spanning intermediate to olympiad-level difficulty. For \textit{out-of-domain} transfer, we evaluate zero-shot on GPQA-Diamond~\citep{Rein2023GPQAAG}, MUSR~\citep{Sprague2023MuSRTT}, and LSAT-AR~\citep{wang2022lsat}, covering scientific QA, multi-step narrative reasoning, and logical reasoning, respectively. All results average five stochastic evaluation runs, with per-run variance in Appendix~\ref{app:stat}.

\paragraph{Metrics and inference.}
We report task accuracy (ACC) and the average number of think tokens (TOK), counted inside the \texttt{<think>}...\texttt{</think>} block. All methods share the same decoding configuration and a maximum completion length of 16{,}384. Correctness is judged only from the answer in \texttt{\textbackslash boxed\{\}}, and an abstention receives no credit while remaining in the denominator, so every query is scored regardless of whether the method chose to answer it.

\begin{table*}[t]
\centering
\small
\setlength{\tabcolsep}{3.2pt}
\renewcommand{\arraystretch}{1.08}
\vspace{-0.26in}
\caption{\textbf{Main results on seven reasoning benchmarks} ($^{\dagger}$in-domain mathematical, $^{\ddagger}$out-of-domain transfer). $\Delta$ on the Avg.\ columns is relative to the base model. Row colors: \legendbox{c1bg} preflight routing, \legendbox{c2bg} in-generation intervention, \legendbox{c3bg} offline adaptation, \legendbox{c4bg} online post-training. Qwen3-4B covers all four families; the DeepSeek backbones report the online family (all families on all backbones in Tab.~\ref{tab:app_full}).}
\label{tab:main_math}\label{tab:ood_transfer}
\resizebox{\textwidth}{!}{%
\begin{tabular}{l|cccc c ccc c|cccc c ccc c}
\toprule
\multirow{3}{*}{\textbf{Model}} & \multicolumn{9}{c|}{\textbf{Accuracy (\%)} ($\uparrow$)} & \multicolumn{9}{c}{\textbf{\#Tokens (K)} ($\downarrow$)} \\
\cmidrule(lr){2-10}\cmidrule(lr){11-19}
 & \multicolumn{4}{c}{Math$^{\dagger}$} & \multirow{2}{*}{\textbf{Avg.}} & \multicolumn{3}{c}{General$^{\ddagger}$} & \multirow{2}{*}{\textbf{Avg.}} & \multicolumn{4}{c}{Math$^{\dagger}$} & \multirow{2}{*}{\textbf{Avg.}} & \multicolumn{3}{c}{General$^{\ddagger}$} & \multirow{2}{*}{\textbf{Avg.}} \\
\cmidrule(lr){2-5}\cmidrule(lr){7-9}\cmidrule(lr){11-14}\cmidrule(lr){16-18}
 & Omni & MATH & AMC & AIME &  & GPQA & MUSR & LSAT &  & Omni & MATH & AMC & AIME &  & GPQA & MUSR & LSAT &  \\
\midrule
\textbf{Qwen3-4B} & 54.5 & 88.8 & 88.0 & 63.3 & 73.6 & 50.9 & 64.4 & 71.7 & 62.4 & 8.6 & 6.8 & 12.9 & 13.8 & 10.5 & 6.5 & 6.1 & 7.2 & 6.6 \\
\rowcolor{c1bg}\quad +DiffAdapt & 48.0 & 84.6 & 84.5 & 63.3 & 70.1 & 48.6 & 53.8 & 48.1 & 50.2 & 6.8 & 6.0 & 10.3 & 11.0 & 8.5 & 6.0 & 6.8 & \underline{3.8} & 5.5 \\
\rowcolor{c2bg}\quad +DEER & 55.3 & 89.4 & \underline{90.5} & \underline{65.3} & \underline{75.1} & \underline{53.2} & 62.0 & 72.1 & 62.4 & 8.1 & 5.2 & 9.7 & 10.3 & 8.3 & 4.8 & 5.4 & 6.2 & 5.4 \\
\rowcolor{c2bg}\quad +KWTQ & 41.4 & 87.8 & 86.0 & 54.0 & 67.3 & 32.1 & 57.5 & 67.1 & 52.2 & 5.4 & 6.7 & 12.6 & 10.3 & 8.8 & \textbf{3.8} & 5.3 & 6.9 & 5.3 \\
\rowcolor{c3bg}\quad +VeriThinker & 49.4 & 86.2 & 86.0 & 63.3 & 71.2 & 50.1 & 47.6 & 59.8 & 52.5 & 6.3 & 5.8 & 9.3 & 10.6 & 8.0 & 6.0 & 3.9 & 4.5 & 4.8 \\
\rowcolor{c3bg}\quad +Overthink & 55.2 & \underline{90.1} & 86.5 & 64.7 & 74.1 & 48.8 & 63.8 & 72.3 & 61.6 & 8.2 & 5.1 & 9.4 & 13.5 & 9.1 & 7.6 & 4.1 & 8.5 & 6.7 \\
\rowcolor{c4bg}\quad +Length-Penalty & 52.9 & 87.4 & 85.5 & 60.0 & 71.4 & 49.5 & 64.2 & 69.2 & 61.0 & 6.6 & 5.4 & 7.2 & 10.9 & 7.5 & 4.7 & 2.5 & 4.5 & 3.9 \\
\rowcolor{c4bg}\quad +CODA & 55.0 & 89.1 & 88.5 & 64.7 & 74.3 & 51.2 & \underline{64.5} & 73.0 & 62.9 & 6.9 & 5.1 & 9.8 & 11.8 & 8.4 & 5.5 & 3.9 & 4.9 & 4.8 \\
\rowcolor{c4bg}\quad +DR.SAF & 54.1 & 89.8 & 90.0 & 64.0 & 74.5 & 50.7 & 63.4 & 72.0 & 62.0 & 5.1 & 4.4 & 6.6 & 9.1 & 6.3 & \underline{4.2} & 2.1 & 3.9 & \underline{3.4} \\
\rowcolor{betbg}\quad +\textbf{\BET}~{\scriptsize($\delta$=0.10)} & \underline{55.6} & 90.0 & 89.5 & 64.7 & 74.9\,{\dsz\textcolor{dteal}{($+$1.3)}} & 52.4 & 63.9 & \underline{74.3} & \underline{63.5}\,{\dsz\textcolor{dteal}{($+$1.2)}} & \textbf{3.8} & \textbf{3.3} & \textbf{5.0} & \textbf{6.1} & \textbf{4.6}\,{\dsz\textcolor{dteal}{($-$57\%)}} & \textbf{3.8} & \textbf{1.2} & \textbf{3.6} & \textbf{2.9}\,{\dsz\textcolor{dteal}{($-$56\%)}} \\
\rowcolor{betbg}\quad +\textbf{\BET}~{\scriptsize($\delta$=0.05)} & \textbf{55.9} & \textbf{90.8} & \textbf{91.0} & \textbf{66.0} & \textbf{75.9}\,{\dsz\textcolor{dteal}{($+$2.3)}} & \textbf{53.9} & \textbf{64.7} & \textbf{75.7} & \textbf{64.8}\,{\dsz\textcolor{dteal}{($+$2.4)}} & \underline{5.0} & \underline{4.2} & \underline{6.1} & \underline{7.7} & \underline{5.8}\,{\dsz\textcolor{dteal}{($-$45\%)}} & 4.6 & \underline{2.0} & 4.5 & 3.7\,{\dsz\textcolor{dteal}{($-$44\%)}} \\
\midrule
\textbf{DeepSeek-7B} & 44.4 & 90.2 & 80.0 & 30.0 & 61.2 & 41.5 & \textbf{44.4} & 30.4 & 38.8 & 8.6 & 3.8 & 11.0 & 10.4 & 8.5 & 4.6 & 5.2 & 6.9 & 5.5 \\
\rowcolor{c4bg}\quad +Length-Penalty & 42.8 & 89.4 & 83.0 & 33.3 & 62.1 & 38.5 & 39.2 & 32.4 & 36.7 & 6.2 & 2.2 & 7.4 & 8.2 & 6.0 & 4.5 & 4.8 & 5.9 & 5.1 \\
\rowcolor{c4bg}\quad +CODA & 45.8 & 90.3 & 84.5 & \underline{34.7} & 63.8 & \underline{41.8} & \underline{44.3} & 34.8 & \underline{40.3} & 6.0 & 2.5 & 7.6 & 7.9 & 6.0 & 3.8 & 4.6 & 5.6 & 4.7 \\
\rowcolor{c4bg}\quad +DR.SAF & 45.2 & 89.6 & \underline{85.0} & \textbf{36.0} & 64.0 & 40.2 & 41.8 & \underline{35.4} & 39.1 & \underline{5.3} & \underline{1.4} & \underline{6.2} & 7.3 & \underline{5.0} & \underline{2.8} & \underline{4.2} & 4.5 & \underline{3.8} \\
\rowcolor{betbg}\quad +\textbf{\BET}~{\scriptsize($\delta$=0.10)} & \underline{46.9} & \underline{91.4} & 84.5 & \underline{34.7} & \underline{64.4}\,{\dsz\textcolor{dteal}{($+$3.2)}} & 41.0 & 43.6 & 34.7 & 39.8\,{\dsz\textcolor{dteal}{($+$1.0)}} & \textbf{3.7} & \textbf{1.2} & \textbf{5.6} & \textbf{5.9} & \textbf{4.1}\,{\dsz\textcolor{dteal}{($-$52\%)}} & \textbf{2.0} & \textbf{2.8} & \textbf{2.1} & \textbf{2.3}\,{\dsz\textcolor{dteal}{($-$58\%)}} \\
\rowcolor{betbg}\quad +\textbf{\BET}~{\scriptsize($\delta$=0.05)} & \textbf{47.1} & \textbf{92.0} & \textbf{85.5} & \textbf{36.0} & \textbf{65.2}\,{\dsz\textcolor{dteal}{($+$4.0)}} & \textbf{42.0} & \underline{44.3} & \textbf{35.6} & \textbf{40.6}\,{\dsz\textcolor{dteal}{($+$1.9)}} & \underline{5.3} & 1.8 & 6.4 & \underline{7.0} & 5.2\,{\dsz\textcolor{dteal}{($-$39\%)}} & 2.9 & 4.4 & \underline{4.0} & \underline{3.8}\,{\dsz\textcolor{dteal}{($-$32\%)}} \\
\midrule
\textbf{DeepSeek-14B} & 51.3 & 91.3 & 86.0 & 43.3 & 68.0 & 47.4 & 46.2 & 42.8 & 45.5 & 7.4 & 3.6 & 9.8 & 8.7 & 7.4 & 4.3 & 4.6 & 5.9 & 4.9 \\
\rowcolor{c4bg}\quad +Length-Penalty & 49.5 & 90.7 & 88.5 & 42.0 & 67.7 & 45.0 & 42.3 & 47.0 & 44.7 & 5.9 & 2.4 & 6.8 & 7.2 & 5.6 & 3.9 & 3.6 & 4.8 & 4.1 \\
\rowcolor{c4bg}\quad +CODA & 52.7 & 90.7 & 87.0 & 43.3 & 68.4 & \underline{47.7} & 46.4 & 48.5 & 47.5 & 5.6 & 2.2 & 7.1 & 6.9 & 5.5 & 3.4 & \underline{3.4} & 4.3 & 3.7 \\
\rowcolor{c4bg}\quad +DR.SAF & 52.7 & 91.6 & 87.0 & \underline{44.7} & 69.0 & 46.4 & 43.4 & \textbf{49.3} & 46.3 & \underline{5.2} & \underline{1.6} & \underline{5.4} & \underline{5.9} & \underline{4.5} & \underline{2.6} & \underline{3.4} & \underline{3.5} & \underline{3.2} \\
\rowcolor{betbg}\quad +\textbf{\BET}~{\scriptsize($\delta$=0.10)} & \underline{53.8} & \underline{91.9} & \underline{89.0} & 44.0 & \underline{69.7}\,{\dsz\textcolor{dteal}{($+$1.7)}} & 47.4 & \underline{46.5} & \underline{48.9} & \underline{47.6}\,{\dsz\textcolor{dteal}{($+$2.1)}} & \textbf{3.9} & \textbf{1.4} & \textbf{4.9} & \textbf{5.3} & \textbf{3.9}\,{\dsz\textcolor{dteal}{($-$48\%)}} & \textbf{2.0} & \textbf{2.2} & \textbf{2.7} & \textbf{2.3}\,{\dsz\textcolor{dteal}{($-$53\%)}} \\
\rowcolor{betbg}\quad +\textbf{\BET}~{\scriptsize($\delta$=0.05)} & \textbf{54.7} & \textbf{92.5} & \textbf{89.5} & \textbf{46.0} & \textbf{70.7}\,{\dsz\textcolor{dteal}{($+$2.7)}} & \textbf{47.9} & \textbf{47.0} & \textbf{49.3} & \textbf{48.1}\,{\dsz\textcolor{dteal}{($+$2.6)}} & 6.1 & 2.5 & 5.9 & 6.5 & 5.2\,{\dsz\textcolor{dteal}{($-$29\%)}} & 3.3 & 3.9 & 4.0 & 3.7\,{\dsz\textcolor{dteal}{($-$25\%)}} \\
\bottomrule
\end{tabular}%
}
\vspace{-0.5cm}
\end{table*}

\subsection{Main Results}
\label{sec:main_results}

\paragraph{\ding{182} \BET attains the best accuracy-efficiency frontier on every backbone.}
Since \BET cuts budget only where further reasoning has stopped paying off, in-domain accuracy improves on all three backbones at $\delta{=}0.10$ while token cost falls by roughly half. On Qwen3-4B, accuracy rises from 73.6\% to 74.9\% as tokens fall by 57\%, and the two distilled DeepSeek models gain 3.2 and 1.7 points at 52\% and 48\% savings. DR.SAF is the most efficient baseline throughout, yet \BET exceeds its average accuracy on every backbone while spending 14\% to 28\% fewer tokens.

\paragraph{\ding{183} \BET saves by selective compression rather than uniform shortening.}
Baselines that apply pressure without a solvability signal lose accuracy somewhere on the in-domain suite: DiffAdapt drops to 48.0\% on Omni-Math, Length-Penalty to 60.0\% on AIME-25, and KWTQ, whose halting rule fires once an auxiliary value estimate falls below a threshold, forfeits a quarter of the vanilla accuracy on Omni-Math. \BET improves every in-domain benchmark on both Qwen3-4B and DeepSeek-7B at roughly half the token cost, a combination no baseline achieves, with its largest gain on the AIME-25 split of DeepSeek-7B, where vanilla is weakest (30.0\% to 34.7\%).

\paragraph{\ding{184} \BET exposes a single accuracy-efficiency knob through $\delta$.}
The fold reward $\delta$ sets how readily \BET abstains on low-solvability queries. Abstention earns $+\delta$ only in groups without a success, so raising $\delta$ extends folding from the least solvable queries upward. At $\delta{=}0.05$, \BET holds the highest in-domain average on every backbone and the best AIME-25 accuracy of 66.0\%, while still removing 29\% to 45\% of the tokens. Raising $\delta$ to 0.10 trades about a point of accuracy for a further fifth to a quarter of its remaining tokens, and even then Qwen3-4B matches the most accurate baseline, DEER, at 74.9\% against 75.1\%, on 45\% fewer tokens. \BET therefore defines a tunable policy family, and Appendix~\ref{app:sensitivity} reports the sweep over $\delta$ together with stability across $\beta$ and $p$.

\subsection{Zero-Shot Transfer to Out-of-Domain Reasoning}
\label{sec:transfer}

\paragraph{Learned allocation transfers without adaptation.}
Trained only on mathematics, \BET keeps its accuracy-efficiency profile out of domain. At $\delta{=}0.05$ it attains the highest general average on every backbone, and at $\delta{=}0.10$ it still improves while removing 53\% to 58\% of the vanilla budget. The margin is widest on LSAT-AR, with DeepSeek-14B rising from 42.8\% to 48.9\%, while on MUSR \BET stays within a point of vanilla at a fifth to half of its cost. CODA comes closest, edging past $\delta{=}0.10$ on DeepSeek-7B at twice the tokens, and falls behind at $\delta{=}0.05$. Signals fitted in domain fare worse: DiffAdapt and VeriThinker surrender a sixth to a fifth of the vanilla average on Qwen3-4B, and KWTQ, whose value probe is trained per backbone, retains barely a third on DeepSeek-7B. \BET instead learns its criterion from the policy's own rollouts, so the criterion moves with the policy rather than with the domain it was fitted to.

\begin{figure*}[t]
    \centering
    \includegraphics[width=\textwidth]{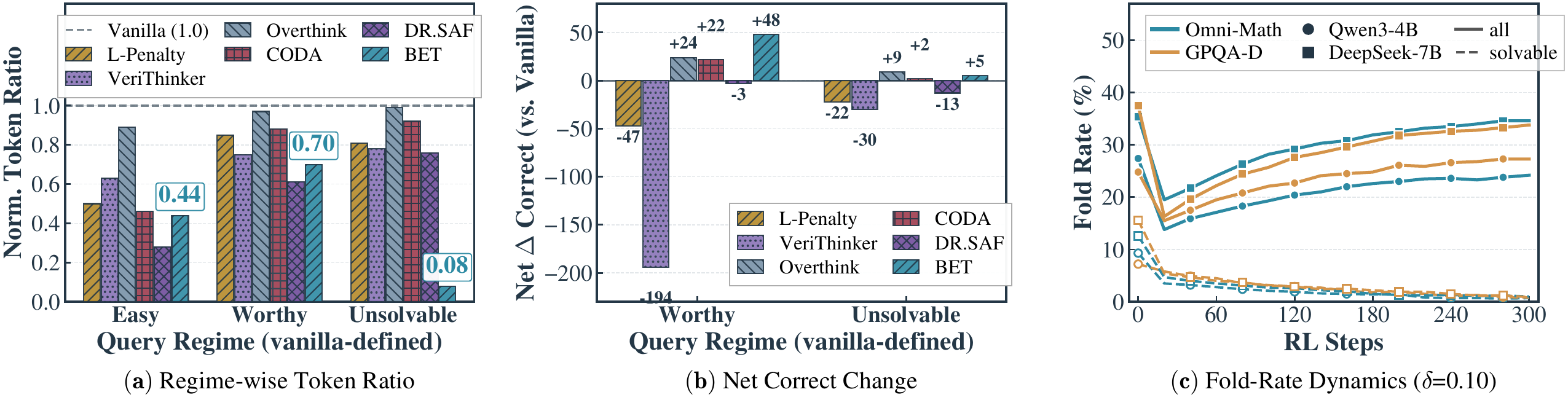}
    \vspace{-0.15in}
\caption{\textbf{Where \BET reallocates compute.} \textbf{(a)} Per-regime token ratio and \textbf{(b)} net correctness change on Omni-Math under the vanilla-derived partition in \S~\ref{sec:emergence}. \textbf{(c)} Fold rate over RL steps on Omni-Math and GPQA-Diamond, for all queries and for those solved at least once by vanilla.}
    \vspace{-0.5cm}
    \label{fig:fig4}
\end{figure*}

\subsection{Where \BET Reallocates Compute}
\label{sec:regime}

We analyze per-regime behavior on Omni-Math under the fixed vanilla-derived partition from \S~\ref{sec:emergence}, comparing token allocation and correctness changes.

\paragraph{\ding{182} \BET induces the clearest solvability-aware allocation profile.}
Most baselines compress nearly uniformly (Fig.~\ref{fig:fig4}(a)): the two offline methods vary by at most 0.15 across regimes, and Length-Penalty, despite compressing easy queries to 0.50, still spends 0.81 of the vanilla budget on unsolvable ones. DR.SAF and CODA separate easy queries more clearly, at 0.28 and 0.46, yet keep 0.76 and 0.92 of it on unsolvable cases, never isolating the zero-return regime. In contrast, \BET matches the target structure, compressing easy queries to 0.44, preserving 0.70 on worthy ones, and collapsing unsolvable queries to 0.08. This profile directly reflects the \textit{short solve}, \textit{hero call}, and \textit{nice fold} behavior induced by the asymmetric reward.

\paragraph{\ding{183} \BET reduces waste and preserves solvable depth.}
Compression without a solvability signal takes part of its savings from reasoning that worthy queries still need, at the cost of lost answers (Fig.~\ref{fig:fig4}(b)). Length-Penalty and DR.SAF lose 47 and 3 worthy instances at budgets of 0.85 and 0.61, whereas \BET gains 48 at 0.70. On vanilla zero queries, explicit fold modeling removes most futile exploration while recovering 5 answers, indicating selective folding rather than blanket refusal.

\paragraph{\ding{184} \BET converges to a selective and stable fold.}
Fig.~\ref{fig:fig4}(c) tracks the fold rate over RL steps on one in-domain and one out-of-domain benchmark, for both backbones. The cold-start policy already emits folds, but not yet tied to solvability, since Stage~1 exposes the action without optimizing its use. As Stage~2 proceeds, the overall rate levels off at a dataset-dependent level, while on queries vanilla solves at least once it falls from 9.3\% to 0.7\% on Omni-Math for Qwen3-4B, and from 12.6\% to 1.0\% on DeepSeek-7B. This separation holds on both benchmarks and backbones. Stage~2 shapes the fold decision itself, and the resulting rate follows each dataset's solvability composition.

\begin{figure*}[t]
    \centering
    \includegraphics[width=\textwidth]{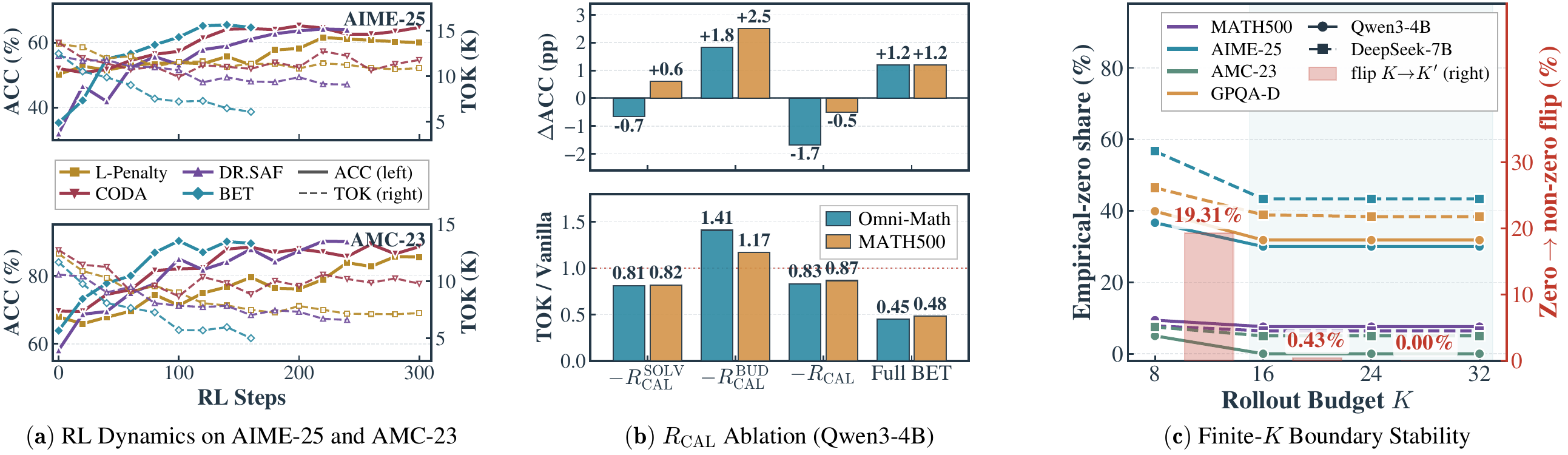}
    \vspace{-0.15in}
\caption{\textbf{Training dynamics, boundary reliability, and calibration ablation.} \textbf{(a)} Accuracy and think tokens per checkpoint, ending at convergence. \textbf{(b)} Ablation of the two calibration branches of $R_{\textsc{cal}}$ on Qwen3-4B. \textbf{(c)} Empirical-zero share and zero-to-non-zero flip rate under nested rollout budgets on MATH500, AIME-25, AMC-23, and GPQA-Diamond with Qwen3-4B and DeepSeek-7B.}
    \vspace{-0.5cm}
    \label{fig:fig5}
\end{figure*}

\subsection{Training Dynamics of \BET}
\label{sec:training_efficiency}

We compare \BET with Length-Penalty, CODA, and DR.SAF on AIME-25 and AMC-23 throughout RL training; Fig.~\ref{fig:fig5}(a) reports accuracy and think-token cost per checkpoint.

\paragraph{\BET reaches its accuracy earlier and at a lower token cost.}
\BET compresses only the reasoning its own rollouts show to be redundant, so accuracy and token cost improve together during training. The cold-start checkpoint starts far below the base model, at 35.3\% and 63.9\% against 63.3\% and 88.0\%, so these gains come from Stage~2. On AIME-25, \BET settles at its final accuracy by step 160, while DR.SAF needs 240 steps and CODA and Length-Penalty the full 300, an ordering AMC-23 preserves. CODA eventually matches \BET's 64.7\% on AIME-25, but only at the end of training and at nearly twice the tokens per query. By step 160, \BET has already established a stable advantage, halving its token cost from 12.5K to 6.1K, the AIME-25 cost it retains at evaluation in Table~\ref{tab:main_math}.

\subsection{Ablation and Boundary Reliability}
\label{sec:deployment}

\paragraph{Solvability calibration is necessary for differentiated allocation.}
Fig.~\ref{fig:fig5}(b) ablates the two branches of $R_{\textsc{cal}}$ on Omni-Math and MATH500. Removing $R_{\textsc{cal}}^{\textsc{solv}}$ weakens partitioning, leaving near-uniform compression at 0.81 and 0.82 of vanilla and Omni-Math accuracy 0.7 points below it. Removing $R_{\textsc{cal}}^{\textsc{bud}}$ detaches planned budget from cost, raising Omni-Math consumption to 1.41 times vanilla for 0.7 points over full \BET. Removing both collapses \BET into undifferentiated compression, dropping below vanilla accuracy at 0.83 of its budget. Full \BET restores the best trade-off at 0.45 and 0.48 of vanilla, so the \texttt{<predict>} field becomes a planning signal only under $R_{\textsc{cal}}$.

\paragraph{A rollout budget of 16 suffices to fix the boundary.}
Since $\hat{s}(x)$ comes from finitely many rollouts, a query enters the zero-return regime only if none of its $K$ samples succeeds, which a small budget can produce by chance. We therefore re-estimate it on four benchmarks and two backbones under strictly nested budgets from $K{=}8$ to $K{=}32$, where each larger budget extends the same prefix and a label can only leave the zero set. The conditional rate at which a zero label acquires a success falls from 19.31\% between 8 and 16 rollouts to 0.43\% between 16 and 24, with no further flips observed at $K{=}32$ (Fig.~\ref{fig:fig5}(c)). Rollout cost grows linearly in $K$ while the correction it buys decays by an order of magnitude, so we adopt $K{=}16$ as the operating point. $K$ thus decides whether success evidence is observed, and the asymmetric gate of Eq.~\ref{eq:rval} decides how abstention is scored once it is.
 
\section{Conclusion}
\label{sec:conclusion}
Cost-efficient reasoning methods control compute without modeling policy-dependent solvability, so they spend budget beyond the model's capability while under-investing in hard-but-solvable queries. We propose \BET, a solvability-aware framework that treats test-time compute as an investment under uncertainty. By combining behavioral cold-start with investment-cost-aware GRPO, \BET aligns solve-or-fold decisions with rollout-derived solvability and induces \textit{short solve}, \textit{nice fold}, and \textit{hero call} behavior. Across seven benchmarks and three base models, it cuts reasoning tokens by 54\% on average while improving accuracy, and transfers zero-shot to scientific and logical reasoning.

\section*{Reproducibility Statement}
The training pipeline, reward formulation, and evaluation protocol needed to reproduce our results are described in \S~\ref{sec:reward}, \S~\ref{sec:training}, and \S~\ref{sec:eval}. The appendices give the complete hyperparameter, hardware, and serving configuration for both stages (Appendix~\ref{app:hyperparams}), the offline profiling procedure and output template behind the cold-start data (Appendix~\ref{app:sft}), the full algorithm (Appendix~\ref{app:algorithm}), every backbone and baseline behind Table~\ref{tab:main_math} (Appendix~\ref{app:full_results}), and the per-run variance of the five evaluation runs that all reported numbers average (Appendix~\ref{app:stat}).
 
\section*{AI Use Statement}
We used generative AI tools as a writing and presentation aid, to edit and polish prose written by the authors and to prepare the scripts that render our measured results into figures and tables. The authors reviewed all AI-assisted output and take responsibility for this paper.
 
\bibliographystyle{iclr2027_conference}
\bibliography{references}

\appendix
\raggedbottom

\clearpage

\section{Training Pipeline and Implementation Details}
\label{app:hyperparams}

\subsection{Overview of the Training Pipeline}
\label{app:training_overview}

\BET is trained with a two-stage pipeline. Stage~1 performs supervised fine-tuning on a compact cold-start dataset constructed from offline profiling of the base model. This stage establishes the structured output format and the three target behaviors used throughout training and evaluation. Stage~2 further optimizes the policy with GRPO under the composite reward defined in \S~\ref{sec:reward}.

Both stages use the same stratified 20K subset of DeepMath-103K~\citep{He2025DeepMath103KAL}, sampled uniformly across difficulty labels. Stage~1 initializes the response format and behavioral repertoire, while Stage~2 learns the corresponding selection policy by recomputing the policy-dependent statistics $\hat{s}(x)$ and $\hat{c}^{*}(x)$ online at each optimization step.

Tables~\ref{tab:app_config}--\ref{tab:aux_symbols} summarize the full configuration used in our experiments. Table~\ref{tab:app_config} reports the Stage~1 supervised fine-tuning hyperparameters alongside the rollout, optimization, decoding, and serving configuration for Stage~2 GRPO training. Table~\ref{tab:app_reward_hparams} lists the reward-related hyperparameters introduced in \S~\ref{sec:reward}. Table~\ref{tab:aux_symbols} defines the auxiliary quantities used by $R_{\textsc{cal}}$, including the calibration targets and the asymmetric budget loss.

\subsection{Training Configuration}
\label{app:sft_hparams}

\paragraph{Implementation notes.}
Stage~1 uses LoRA-based supervised fine-tuning. Stage~2 uses a trainer--server architecture in which policy optimization and rollout generation run asynchronously. We use an 8$\times$A40 training node together with a single RTX~PRO~6000 vLLM server for rollout generation. Unless otherwise stated, the maximum completion length is 16{,}384 tokens, and each prompt in Stage~2 is sampled with $K{=}16$ rollouts. \BET uses the same structured output template throughout Stage~2 training and evaluation.

\begin{table}[H]
\centering
\footnotesize
\setlength{\tabcolsep}{4pt}
\renewcommand{\arraystretch}{1.02}
\caption{Training configuration shared by all base models. Left: Stage~1 supervised fine-tuning. Right: Stage~2 GRPO training, rollout, and evaluation.}
\label{tab:app_config}
\begin{minipage}[t]{0.47\linewidth}
\centering
\begin{tabular}{@{}lr@{}}
\toprule
\textbf{Stage~1 (SFT)} & \textbf{Value} \\
\midrule
Training data & 20K DeepMath-103K \\
Objective & SFT on cold-start data \\
PEFT method & LoRA \\
LoRA target modules & Q, K, V, O \\
LoRA rank $r$ & 64 \\
LoRA scaling $\alpha_{\mathrm{LoRA}}$ & 128 \\
LoRA dropout & 0.05 \\
Optimizer & AdamW \\
Learning rate & $2\times 10^{-5}$ \\
Weight decay & 0.01 \\
Scheduler & Cosine \\
Warmup ratio & 0.03 \\
Epochs & 2 \\
Micro-batch size & 1 \\
Gradient accumulation & 16 \\
Effective batch size & 128 \\
Max sequence length & 16{,}384 \\
Validation split & 5\% \\
Checkpoint selection & Lowest val.\ loss \\
\bottomrule
\end{tabular}
\end{minipage}\hfill
\begin{minipage}[t]{0.51\linewidth}
\centering
\begin{tabular}{@{}lr@{}}
\toprule
\textbf{Stage~2 (GRPO)} & \textbf{Value} \\
\midrule
Initialization & Stage~1 checkpoint \\
RL objective & GRPO (\S~\ref{sec:reward}) \\
Trainer framework & TRL + vLLM server \\
Training GPUs & 8$\times$A40 \\
Inference server & 1$\times$RTX~PRO~6000 \\
Rollouts per query $K$ & 16 \\
Prompt batch size & 64 queries \\
Trajectories per step & 1{,}024 \\
Rollout temperature & 0.8 \\
Rollout top-$p$ & 1.0 \\
Max completion length & 16{,}384 \\
Optimizer & AdamW \\
Learning rate & $1\times 10^{-6}$ \\
Weight decay & 0.0 \\
GRPO clip $\epsilon_{\mathrm{clip}}$ & 0.0625 \\
KL coefficient & 0.0 \\
Training steps & 300 \\
Evaluation cadence & Every 20 steps \\
Evaluation decoding & $T{=}0.8$, top-$p{=}1.0$ \\
Evaluation repeats & 5 per query \\
Output template & App.~\ref{app:sft_format} \\
Format checker & Enforced \\
\bottomrule
\end{tabular}
\end{minipage}
\end{table}

\subsection{Reward Hyperparameters and Auxiliary Definitions}
\label{app:reward_hparams}\label{app:aux_symbols}

\begin{table}[H]
\centering
\footnotesize
\setlength{\tabcolsep}{4pt}
\renewcommand{\arraystretch}{1.06}
\caption{Reward-related hyperparameters used in \S~\ref{sec:reward}.}
\label{tab:app_reward_hparams}
\begin{tabularx}{\linewidth}{p{0.12\linewidth} p{0.24\linewidth} p{0.09\linewidth} Y}
\toprule
\textbf{Symbol} & \textbf{Meaning} & \textbf{Value} & \textbf{Role in BET} \\
\midrule
$\delta$ & Reward for correct abstention & 0.10 & Encourages Nice Fold on effectively unsolvable queries \\
$\lambda$ & Penalty for premature abstention & 0.80 & Discourages folding on queries that warrant deeper investment \\
$\beta$ & Efficiency bonus strength & 0.30 & Rewards Short Solve when a concise correct path exists \\
$\alpha_{\mathrm{fail}}$ & Failed-attempt cost weight & 0.20 & Scales the length-normalized penalty $\phi(c)=\alpha_{\mathrm{fail}}c/L_{\max}$ on incorrect attempts \\
$\tau$ & Solvability confidence threshold & 0.20 & Activates $R_{\textsc{eff}}$ only when solvability is reliable \\
$\epsilon_{\mathrm{abs}}$ & Abstention guard threshold & $1/K$ & Makes fold rewarding only when all $K$ rollouts fail \\
$p$ & Efficient-cost percentile & 30\% & Defines the lower-envelope estimate of $\hat{c}^{*}(x)$ \\
$\gamma_s$ & Solvability calibration weight & 0.10 & Aligns predicted solvability with empirical group solvability \\
$\gamma_b$ & Budget calibration weight & 0.20 & Aligns requested budget with efficient solution cost \\
$\gamma'_s$ & Zero-return solvability weight & 0.20 & Sharpens solvability calibration in the zero-return regime \\
$\gamma'_b$ & Unsolvable-regime budget weight & 0.10 & Suppresses over-budgeting when Nice Fold is preferred \\
$\mu$ & Underestimation penalty multiplier & 2.0 & Makes under-budgeting costlier than over-budgeting in $\ell(\hat{b}, b^{*})$ \\
\bottomrule
\end{tabularx}
\end{table}

Table~\ref{tab:aux_symbols} provides the definitions of the calibration targets and the asymmetric budget loss used in Eq.~\eqref{eq:rcal}. In the solvable regime, $R_{\textsc{cal}}$ aligns the model's predicted solvability $\hat{s}$ directly with the group estimate $\hat{s}(x)$. The budget target $b^{*}(x)$ normalizes the efficient solution cost by the maximum completion length $L_{\max}$. The asymmetric loss $\ell(\hat{b}, b^{*})$ penalizes budget underestimation more heavily than overestimation through the coefficient $\mu > 1$, reflecting the observation that under-budgeting truncates productive reasoning while over-budgeting only leaves unused headroom.

\begin{table}[H]
\centering
\small
\setlength{\tabcolsep}{4pt}
\renewcommand{\arraystretch}{1.06}
\caption{Auxiliary symbols in the calibration reward $R_{\textsc{cal}}$ (Eq.~\ref{eq:rcal}).}
\label{tab:aux_symbols}
\begin{tabularx}{\linewidth}{p{0.16\linewidth} p{0.36\linewidth} Y}
\toprule
\textbf{Symbol} & \textbf{Definition} & \textbf{Role} \\
\midrule

$b^{*}(x)$ 
& $\hat{c}^{*}(x) / L_{\max}$ 
& Budget target expressing efficient cost as a context fraction \\

$\ell(\hat{b}, b^{*})$ 
& $\begin{cases}
\mu |\hat{b} - b^{*}|, & \hat{b} < b^{*} \\
|\hat{b} - b^{*}|, & \hat{b} \geq b^{*}
\end{cases}$ 
& Asymmetric budget loss with heavier penalty on underestimation \\
\bottomrule
\end{tabularx}
\end{table}

\subsection{Baseline Implementation}
\label{app:baselines}

The eight baselines are grouped by the stage at which they control compute. \textit{Preflight routing} (DiffAdapt) selects a reasoning mode before decoding begins. \textit{In-generation intervention} (DEER, KWTQ) halts decoding on a runtime signal. DEER exits once answer confidence stabilizes, and KWTQ abstains once an auxiliary value estimate falls below its threshold, with that estimator trained separately for each backbone. \textit{Offline adaptation} (VeriThinker, Overthink) fine-tunes the backbone on its published supervision and then decodes without further intervention. \textit{Online post-training} (Length-Penalty, CODA, DR.SAF) starts from the same base checkpoint as \BET and uses the same rollout group size $K{=}16$, decoding configuration, maximum completion length, training steps, and evaluation protocol. Method-specific coefficients follow the original papers when specified, and are otherwise chosen by a small validation sweep on accuracy and think-token count, then held fixed across all evaluation benchmarks. Every method shares the same external prompt, answer extraction, and scoring rule, and is evaluated under the same five stochastic runs.

\section{Complete Main Results}
\label{app:full_results}
 
Table~\ref{tab:main_math} in the main text reports all four method families on Qwen3-4B and the online post-training family on the two DeepSeek backbones.  Table~\ref{tab:app_full} below completes that comparison: every backbone carries all eight baselines across the four families, on all seven benchmarks.  Numbers, formatting, and row colors follow Table~\ref{tab:main_math}, and every entry averages the same five stochastic evaluation runs.  With all baselines included, \BET at $\delta{=}0.05$ still holds the best in-domain and general average on all three backbones, and among the methods that improve on the base model's accuracy, \BET at $\delta{=}0.10$ spends the fewest tokens on every backbone.
 
\begin{table}[H]
\centering
\small
\setlength{\tabcolsep}{3.2pt}
\renewcommand{\arraystretch}{1.05}
\caption{\textbf{Complete results on seven reasoning benchmarks for all three backbones} ($^{\dagger}$in-domain mathematical, $^{\ddagger}$out-of-domain transfer). This table extends Tab.~\ref{tab:main_math}, which reports the four method families on Qwen3-4B and the online family on the two DeepSeek backbones. $\Delta$ on the Avg.\ columns is relative to the base model. Row colors: \legendbox{c1bg} preflight routing, \legendbox{c2bg} in-generation intervention, \legendbox{c3bg} offline adaptation, \legendbox{c4bg} online post-training.}
\label{tab:app_full}
\resizebox{\textwidth}{!}{%
\begin{tabular}{l|cccc c ccc c|cccc c ccc c}
\toprule
\multirow{3}{*}{\textbf{Model}} & \multicolumn{9}{c|}{\textbf{Accuracy (\%)} ($\uparrow$)} & \multicolumn{9}{c}{\textbf{\#Tokens (K)} ($\downarrow$)} \\
\cmidrule(lr){2-10}\cmidrule(lr){11-19}
 & \multicolumn{4}{c}{Math$^{\dagger}$} & \multirow{2}{*}{\textbf{Avg.}} & \multicolumn{3}{c}{General$^{\ddagger}$} & \multirow{2}{*}{\textbf{Avg.}} & \multicolumn{4}{c}{Math$^{\dagger}$} & \multirow{2}{*}{\textbf{Avg.}} & \multicolumn{3}{c}{General$^{\ddagger}$} & \multirow{2}{*}{\textbf{Avg.}} \\
\cmidrule(lr){2-5}\cmidrule(lr){7-9}\cmidrule(lr){11-14}\cmidrule(lr){16-18}
 & Omni & MATH & AMC & AIME &  & GPQA & MUSR & LSAT &  & Omni & MATH & AMC & AIME &  & GPQA & MUSR & LSAT &  \\
\midrule
\textbf{Qwen3-4B} & 54.5 & 88.8 & 88.0 & 63.3 & 73.6 & 50.9 & 64.4 & 71.7 & 62.4 & 8.6 & 6.8 & 12.9 & 13.8 & 10.5 & 6.5 & 6.1 & 7.2 & 6.6 \\
\rowcolor{c1bg}\quad +DiffAdapt & 48.0 & 84.6 & 84.5 & 63.3 & 70.1 & 48.6 & 53.8 & 48.1 & 50.2 & 6.8 & 6.0 & 10.3 & 11.0 & 8.5 & 6.0 & 6.8 & \underline{3.8} & 5.5 \\
\rowcolor{c2bg}\quad +DEER & 55.3 & 89.4 & \underline{90.5} & \underline{65.3} & \underline{75.1} & \underline{53.2} & 62.0 & 72.1 & 62.4 & 8.1 & 5.2 & 9.7 & 10.3 & 8.3 & 4.8 & 5.4 & 6.2 & 5.4 \\
\rowcolor{c2bg}\quad +KWTQ & 41.4 & 87.8 & 86.0 & 54.0 & 67.3 & 32.1 & 57.5 & 67.1 & 52.2 & 5.4 & 6.7 & 12.6 & 10.3 & 8.8 & \textbf{3.8} & 5.3 & 6.9 & 5.3 \\
\rowcolor{c3bg}\quad +VeriThinker & 49.4 & 86.2 & 86.0 & 63.3 & 71.2 & 50.1 & 47.6 & 59.8 & 52.5 & 6.3 & 5.8 & 9.3 & 10.6 & 8.0 & 6.0 & 3.9 & 4.5 & 4.8 \\
\rowcolor{c3bg}\quad +Overthink & 55.2 & \underline{90.1} & 86.5 & 64.7 & 74.1 & 48.8 & 63.8 & 72.3 & 61.6 & 8.2 & 5.1 & 9.4 & 13.5 & 9.1 & 7.6 & 4.1 & 8.5 & 6.7 \\
\rowcolor{c4bg}\quad +Length-Penalty & 52.9 & 87.4 & 85.5 & 60.0 & 71.4 & 49.5 & 64.2 & 69.2 & 61.0 & 6.6 & 5.4 & 7.2 & 10.9 & 7.5 & 4.7 & 2.5 & 4.5 & 3.9 \\
\rowcolor{c4bg}\quad +CODA & 55.0 & 89.1 & 88.5 & 64.7 & 74.3 & 51.2 & \underline{64.5} & 73.0 & 62.9 & 6.9 & 5.1 & 9.8 & 11.8 & 8.4 & 5.5 & 3.9 & 4.9 & 4.8 \\
\rowcolor{c4bg}\quad +DR.SAF & 54.1 & 89.8 & 90.0 & 64.0 & 74.5 & 50.7 & 63.4 & 72.0 & 62.0 & 5.1 & 4.4 & 6.6 & 9.1 & 6.3 & \underline{4.2} & 2.1 & 3.9 & \underline{3.4} \\
\rowcolor{betbg}\quad +\textbf{\BET}~{\scriptsize($\delta$=0.10)} & \underline{55.6} & 90.0 & 89.5 & 64.7 & 74.9\,{\dsz\textcolor{dteal}{($+$1.3)}} & 52.4 & 63.9 & \underline{74.3} & \underline{63.5}\,{\dsz\textcolor{dteal}{($+$1.2)}} & \textbf{3.8} & \textbf{3.3} & \textbf{5.0} & \textbf{6.1} & \textbf{4.6}\,{\dsz\textcolor{dteal}{($-$57\%)}} & \textbf{3.8} & \textbf{1.2} & \textbf{3.6} & \textbf{2.9}\,{\dsz\textcolor{dteal}{($-$56\%)}} \\
\rowcolor{betbg}\quad +\textbf{\BET}~{\scriptsize($\delta$=0.05)} & \textbf{55.9} & \textbf{90.8} & \textbf{91.0} & \textbf{66.0} & \textbf{75.9}\,{\dsz\textcolor{dteal}{($+$2.3)}} & \textbf{53.9} & \textbf{64.7} & \textbf{75.7} & \textbf{64.8}\,{\dsz\textcolor{dteal}{($+$2.4)}} & \underline{5.0} & \underline{4.2} & \underline{6.1} & \underline{7.7} & \underline{5.8}\,{\dsz\textcolor{dteal}{($-$45\%)}} & 4.6 & \underline{2.0} & 4.5 & 3.7\,{\dsz\textcolor{dteal}{($-$44\%)}} \\
\midrule
\textbf{DeepSeek-7B} & 44.4 & 90.2 & 80.0 & 30.0 & 61.2 & 41.5 & \textbf{44.4} & 30.4 & 38.8 & 8.6 & 3.8 & 11.0 & 10.4 & 8.5 & 4.6 & 5.2 & 6.9 & 5.5 \\
\rowcolor{c1bg}\quad +DiffAdapt & 42.0 & 90.8 & 80.0 & 30.7 & 60.9 & 37.4 & 36.2 & 23.4 & 32.3 & 9.0 & 3.0 & 11.4 & 9.7 & 8.3 & 4.0 & 5.7 & 5.0 & 4.9 \\
\rowcolor{c2bg}\quad +DEER & 43.4 & 89.0 & 74.5 & 33.3 & 60.1 & 41.7 & 34.4 & 28.7 & 34.9 & 7.3 & 2.8 & 8.4 & 8.3 & 6.7 & 3.8 & 4.0 & 5.6 & 4.5 \\
\rowcolor{c2bg}\quad +KWTQ & 29.9 & 88.7 & 76.5 & 14.7 & 52.4 & 16.7 & 20.9 & 4.3 & 13.9 & \underline{4.1} & 3.7 & 10.2 & \textbf{3.8} & 5.5 & \textbf{1.7} & \textbf{2.1} & \textbf{1.7} & \textbf{1.8} \\
\rowcolor{c3bg}\quad +VeriThinker & 45.2 & \textbf{92.2} & 83.0 & 34.0 & 63.6 & 40.2 & 29.8 & 30.6 & 33.5 & 7.8 & 2.4 & 9.2 & 8.6 & 7.0 & 3.9 & 4.9 & 6.2 & 5.0 \\
\rowcolor{c3bg}\quad +Overthink & 43.9 & 91.6 & 82.5 & 33.3 & 62.8 & 40.9 & 39.0 & 29.6 & 36.5 & 8.1 & 2.6 & 9.3 & 9.4 & 7.3 & 4.4 & 5.0 & 6.5 & 5.3 \\
\rowcolor{c4bg}\quad +Length-Penalty & 42.8 & 89.4 & 83.0 & 33.3 & 62.1 & 38.5 & 39.2 & 32.4 & 36.7 & 6.2 & 2.2 & 7.4 & 8.2 & 6.0 & 4.5 & 4.8 & 5.9 & 5.1 \\
\rowcolor{c4bg}\quad +CODA & 45.8 & 90.3 & 84.5 & \underline{34.7} & 63.8 & \underline{41.8} & \underline{44.3} & 34.8 & \underline{40.3} & 6.0 & 2.5 & 7.6 & 7.9 & 6.0 & 3.8 & 4.6 & 5.6 & 4.7 \\
\rowcolor{c4bg}\quad +DR.SAF & 45.2 & 89.6 & \underline{85.0} & \textbf{36.0} & 64.0 & 40.2 & 41.8 & \underline{35.4} & 39.1 & 5.3 & \underline{1.4} & \underline{6.2} & 7.3 & \underline{5.0} & 2.8 & 4.2 & 4.5 & 3.8 \\
\rowcolor{betbg}\quad +\textbf{\BET}~{\scriptsize($\delta$=0.10)} & \underline{46.9} & 91.4 & 84.5 & \underline{34.7} & \underline{64.4}\,{\dsz\textcolor{dteal}{($+$3.2)}} & 41.0 & 43.6 & 34.7 & 39.8\,{\dsz\textcolor{dteal}{($+$1.0)}} & \textbf{3.7} & \textbf{1.2} & \textbf{5.6} & \underline{5.9} & \textbf{4.1}\,{\dsz\textcolor{dteal}{($-$52\%)}} & \underline{2.0} & \underline{2.8} & \underline{2.1} & \underline{2.3}\,{\dsz\textcolor{dteal}{($-$58\%)}} \\
\rowcolor{betbg}\quad +\textbf{\BET}~{\scriptsize($\delta$=0.05)} & \textbf{47.1} & \underline{92.0} & \textbf{85.5} & \textbf{36.0} & \textbf{65.2}\,{\dsz\textcolor{dteal}{($+$4.0)}} & \textbf{42.0} & \underline{44.3} & \textbf{35.6} & \textbf{40.6}\,{\dsz\textcolor{dteal}{($+$1.9)}} & 5.3 & 1.8 & 6.4 & 7.0 & 5.2\,{\dsz\textcolor{dteal}{($-$39\%)}} & 2.9 & 4.4 & 4.0 & 3.8\,{\dsz\textcolor{dteal}{($-$32\%)}} \\
\midrule
\textbf{DeepSeek-14B} & 51.3 & 91.3 & 86.0 & 43.3 & 68.0 & 47.4 & 46.2 & 42.8 & 45.5 & 7.4 & 3.6 & 9.8 & 8.7 & 7.4 & 4.3 & 4.6 & 5.9 & 4.9 \\
\rowcolor{c1bg}\quad +DiffAdapt & 49.2 & 91.7 & 86.0 & 43.3 & 67.6 & 46.4 & 39.1 & 36.0 & 40.5 & 7.3 & 3.0 & 9.7 & 8.0 & 7.0 & 3.9 & 5.0 & 4.3 & 4.4 \\
\rowcolor{c2bg}\quad +DEER & 50.5 & 90.9 & 83.0 & 43.3 & 67.0 & 47.5 & 37.7 & 41.6 & 42.3 & 6.6 & 2.8 & 7.8 & 7.1 & 6.1 & 3.6 & 3.8 & 5.0 & 4.1 \\
\rowcolor{c2bg}\quad +KWTQ & 37.8 & 90.5 & 85.0 & 26.7 & 60.0 & 23.7 & 25.1 & 9.6 & 19.5 & \underline{4.2} & 3.5 & 9.3 & \textbf{4.3} & 5.3 & \textbf{2.0} & \underline{2.5} & \textbf{2.3} & \textbf{2.3} \\
\rowcolor{c3bg}\quad +VeriThinker & 51.9 & \textbf{92.7} & 87.0 & 44.0 & 68.9 & \underline{47.7} & 33.2 & 42.2 & 41.0 & 6.6 & 2.6 & 8.2 & 7.3 & 6.2 & 3.8 & 4.1 & 5.1 & 4.3 \\
\rowcolor{c3bg}\quad +Overthink & 50.9 & 92.2 & 87.0 & 43.3 & 68.4 & 47.5 & 41.9 & 42.4 & 43.9 & 7.1 & 2.7 & 8.3 & 8.1 & 6.5 & 4.3 & 4.2 & 5.8 & 4.8 \\
\rowcolor{c4bg}\quad +Length-Penalty & 49.5 & 90.7 & 88.5 & 42.0 & 67.7 & 45.0 & 42.3 & 47.0 & 44.7 & 5.9 & 2.4 & 6.8 & 7.2 & 5.6 & 3.9 & 3.6 & 4.8 & 4.1 \\
\rowcolor{c4bg}\quad +CODA & 52.7 & 90.7 & 87.0 & 43.3 & 68.4 & \underline{47.7} & 46.4 & 48.5 & 47.5 & 5.6 & 2.2 & 7.1 & 6.9 & 5.5 & 3.4 & 3.4 & 4.3 & 3.7 \\
\rowcolor{c4bg}\quad +DR.SAF & 52.7 & 91.6 & 87.0 & \underline{44.7} & 69.0 & 46.4 & 43.4 & \textbf{49.3} & 46.3 & 5.2 & \underline{1.6} & \underline{5.4} & 5.9 & \underline{4.5} & \underline{2.6} & 3.4 & 3.5 & \underline{3.2} \\
\rowcolor{betbg}\quad +\textbf{\BET}~{\scriptsize($\delta$=0.10)} & \underline{53.8} & 91.9 & \underline{89.0} & 44.0 & \underline{69.7}\,{\dsz\textcolor{dteal}{($+$1.7)}} & 47.4 & \underline{46.5} & \underline{48.9} & \underline{47.6}\,{\dsz\textcolor{dteal}{($+$2.1)}} & \textbf{3.9} & \textbf{1.4} & \textbf{4.9} & \underline{5.3} & \textbf{3.9}\,{\dsz\textcolor{dteal}{($-$48\%)}} & \textbf{2.0} & \textbf{2.2} & \underline{2.7} & \textbf{2.3}\,{\dsz\textcolor{dteal}{($-$53\%)}} \\
\rowcolor{betbg}\quad +\textbf{\BET}~{\scriptsize($\delta$=0.05)} & \textbf{54.7} & \underline{92.5} & \textbf{89.5} & \textbf{46.0} & \textbf{70.7}\,{\dsz\textcolor{dteal}{($+$2.7)}} & \textbf{47.9} & \textbf{47.0} & \textbf{49.3} & \textbf{48.1}\,{\dsz\textcolor{dteal}{($+$2.6)}} & 6.1 & 2.5 & 5.9 & 6.5 & 5.2\,{\dsz\textcolor{dteal}{($-$29\%)}} & 3.3 & 3.9 & 4.0 & 3.7\,{\dsz\textcolor{dteal}{($-$25\%)}} \\
\bottomrule
\end{tabular}%
}
\end{table}

\section{Statistical Significance}
\label{app:stat}
 
Table~\ref{tab:std} reports the mean and standard deviation of accuracy and think-token count over the five stochastic evaluation runs behind Table~\ref{tab:main_math}, on four benchmarks with the Qwen3-4B backbone.  Three are in-domain and one is out-of-domain, and the set includes AMC-23 and AIME-25, the two smallest evaluation sets and those with the largest run-to-run variance.  All runs use the same trained checkpoint with independent stochastic sampling ($T{=}0.8$, top-$p{=}1.0$), and the standard deviation is computed with Bessel's correction ($N{-}1$ denominator) and reported as $1\sigma$.  The means match the corresponding entries of Table~\ref{tab:main_math}.
 
\begin{table}[H]
\centering
\caption{Mean $\pm$ standard deviation ($1\sigma$, Bessel-corrected) over the five stochastic evaluation runs behind Tab.~\ref{tab:main_math}, on the Qwen3-4B backbone.  Accuracy in \% and think tokens in K, matching Tab.~\ref{tab:main_math}. Row colors follow the same four method families.}
\label{tab:std}
\footnotesize
\setlength{\tabcolsep}{3.5pt}
\renewcommand{\arraystretch}{1.08}
\resizebox{\textwidth}{!}{%
\begin{tabular}{l cc cc cc cc}
\toprule
\multirow{2}{*}{\textbf{Model Name}} & \multicolumn{2}{c}{\textbf{Omni-Math}$^{\dagger}$} & \multicolumn{2}{c}{\textbf{AMC-23}$^{\dagger}$} & \multicolumn{2}{c}{\textbf{AIME-25}$^{\dagger}$} & \multicolumn{2}{c}{\textbf{GPQA-Diamond}$^{\ddagger}$} \\
\cmidrule(lr){2-3}\cmidrule(lr){4-5}\cmidrule(lr){6-7}\cmidrule(lr){8-9}
 & Acc.\ (\%) $\uparrow$ & \#Tok.\ (K) $\downarrow$ & Acc.\ (\%) $\uparrow$ & \#Tok.\ (K) $\downarrow$ & Acc.\ (\%) $\uparrow$ & \#Tok.\ (K) $\downarrow$ & Acc.\ (\%) $\uparrow$ & \#Tok.\ (K) $\downarrow$ \\
\midrule
Qwen3-4B-Thinking-2507
  & $54.5 \pm 0.38$ & $8.6 \pm 0.13$ & $88.0 \pm 1.12$ & $12.9 \pm 0.69$ & $63.3 \pm 2.36$ & $13.8 \pm 0.70$ & $50.9 \pm 2.18$ & $6.5 \pm 0.29$ \\
\rowcolor{c1bg}\quad +DiffAdapt
  & $48.0 \pm 0.47$ & $6.8 \pm 0.14$ & $84.5 \pm 2.09$ & $10.3 \pm 0.62$ & $63.3 \pm 2.36$ & $11.0 \pm 0.66$ & $48.6 \pm 2.71$ & $6.0 \pm 0.27$ \\
\rowcolor{c2bg}\quad +DEER
  & $55.3 \pm 0.36$ & $8.1 \pm 0.11$ & $90.5 \pm 1.12$ & $9.7 \pm 0.59$ & $65.3 \pm 1.83$ & $10.3 \pm 0.63$ & $53.2 \pm 2.07$ & $4.8 \pm 0.25$ \\
\rowcolor{c2bg}\quad +KWTQ
  & $41.4 \pm 0.50$ & $5.4 \pm 0.09$ & $86.0 \pm 1.37$ & $12.6 \pm 0.68$ & $54.0 \pm 3.65$ & $10.3 \pm 0.68$ & $32.1 \pm 2.71$ & $3.8 \pm 0.22$ \\
\rowcolor{c3bg}\quad +VeriThinker
  & $49.4 \pm 0.43$ & $6.3 \pm 0.13$ & $86.0 \pm 1.37$ & $9.3 \pm 0.58$ & $63.3 \pm 2.36$ & $10.6 \pm 0.63$ & $50.1 \pm 2.46$ & $6.0 \pm 0.28$ \\
\rowcolor{c3bg}\quad +Overthink
  & $55.2 \pm 0.39$ & $8.2 \pm 0.12$ & $86.5 \pm 1.37$ & $9.4 \pm 0.58$ & $64.7 \pm 1.83$ & $13.5 \pm 0.69$ & $48.8 \pm 2.25$ & $7.6 \pm 0.31$ \\
\rowcolor{c4bg}\quad +Length-Penalty
  & $52.9 \pm 0.48$ & $6.6 \pm 0.15$ & $85.5 \pm 2.09$ & $7.2 \pm 0.50$ & $60.0 \pm 2.36$ & $10.9 \pm 0.65$ & $49.5 \pm 2.77$ & $4.7 \pm 0.24$ \\
\rowcolor{c4bg}\quad +CODA
  & $55.0 \pm 0.41$ & $6.9 \pm 0.11$ & $88.5 \pm 1.37$ & $9.8 \pm 0.60$ & $64.7 \pm 2.98$ & $11.8 \pm 0.73$ & $51.2 \pm 2.38$ & $5.5 \pm 0.26$ \\
\rowcolor{c4bg}\quad +DR.SAF
  & $54.1 \pm 0.41$ & $5.1 \pm 0.10$ & $90.0 \pm 1.77$ & $6.6 \pm 0.47$ & $64.0 \pm 1.49$ & $9.1 \pm 0.57$ & $50.7 \pm 2.36$ & $4.2 \pm 0.23$ \\
\rowcolor{betbg}\quad +\textbf{\BET}~{\scriptsize($\delta$=0.10)}
  & $55.6 \pm 0.34$ & $3.8 \pm 0.08$ & $89.5 \pm 1.12$ & $5.0 \pm 0.40$ & $64.7 \pm 1.83$ & $6.1 \pm 0.43$ & $52.4 \pm 1.97$ & $3.8 \pm 0.22$ \\
\rowcolor{betbg}\quad +\textbf{\BET}~{\scriptsize($\delta$=0.05)}
  & $55.9 \pm 0.36$ & $5.0 \pm 0.09$ & $91.0 \pm 1.37$ & $6.1 \pm 0.46$ & $66.0 \pm 1.49$ & $7.7 \pm 0.44$ & $53.9 \pm 2.06$ & $4.6 \pm 0.24$ \\
\bottomrule
\end{tabular}%
}
\end{table}

\section{Cold-Start Data Construction and Output Format}
\label{app:sft}

\subsection{Offline Profiling and Demonstration Construction}
\label{app:sft_profiling}

The cold-start data in Stage~1 is constructed by offline profiling of the base policy on the same 20K stratified subset of DeepMath-103K used throughout training. Stage~1 is behavioral supervision that exposes the model to the output format and to the three target behaviors. For each query $x$ we draw a single sample $y_x$ from the base policy under the same maximum completion length as in the main experiments, and record whether it is correct together with its think-token cost $c(y_x)$.
 
The policy-dependent statistics that drive the reward are not taken from this pass. The solvability estimate $\hat{s}(x)$ and the efficient solution cost $\hat{c}^{*}(x)$ of \S~\ref{sec:dynamic_estimation} are computed from $K{=}16$ rollouts and are recomputed online at every Stage~2 step from the evolving policy, so \BET learns its decision boundary in Stage~2. Stage~1 supplies format and behavioral coverage, and Stage~2 supplies the criterion.
 
We partition profiled queries into three buckets and construct one demonstration for each selected query:
\begin{itemize}[leftmargin=*]
    \item \textbf{\textit{Short Solve.}} Queries whose sampled solution is correct and inexpensive, taken from the lower-cost portion of the correctly answered set. We retain the sampled trace and assign a low predicted budget.
    \item \textbf{\textit{Hero Call.}} Queries whose sampled solution is correct but expensive, taken from the higher-cost portion of that set. We retain the sampled trace and assign a higher predicted budget, exposing the model to the amount of exploration needed in the high-cost solvable regime.
    \item \textbf{\textit{Nice Fold.}} Queries whose sampled solution is incorrect. For these queries, the target output contains an explicit abstention decision, together with a low predicted budget and no reasoning chain beyond the abstention.
\end{itemize}

In practice, we rank the correctly answered queries by $c(y_x)$ and take the lower and upper portions of this list to form the \textit{short solve} and \textit{hero call} buckets, so Stage~1 teaches distinct budget regimes. \textit{Nice fold} examples are taken from the queries the base policy fails on this pass.  The \texttt{Solvability} and \texttt{Budget} fields of a Stage~1 target carry behavioral values. They are set from a monotone mapping of the query's difficulty label and the cost of its sampled solution, and to zero for nice fold targets, which establishes the semantics of each field and the ordering across the three behaviors. In Stage~2, $R_{\textsc{cal}}$ aligns them with the online $\hat{s}(x)$ and $\hat{c}^{*}(x)$.  Table~\ref{tab:app_coldstart_stats} summarizes the resulting dataset.

\begin{table}[h]
\centering
\small
\setlength{\tabcolsep}{5pt}
\renewcommand{\arraystretch}{1.12}
\caption{Statistics of the Stage~1 cold-start dataset.}
\label{tab:app_coldstart_stats}
\begin{tabular}{lcccc}
\toprule
\textbf{Demonstration type} & \textbf{Selection rule} & \textbf{\# Examples} & \textbf{Avg. think tokens} & \textbf{Avg. total tokens} \\
\midrule
Short Solve & correct sample, low $c(y_x)$ & 7{,}641 & 1093 & 1172 \\
Hero Call & correct sample, high $c(y_x)$ & 5{,}428 & 5524 & 5621 \\
Nice Fold & incorrect sample & 6{,}931 & 38 & 61 \\
\midrule
Total & -- & 20{,}000 & 1930 & 1994 \\
\bottomrule
\end{tabular}
\end{table}
\subsection{Structured Output Template}
\label{app:sft_format}

All Stage~1 demonstrations use the same structured template shown in Fig.~\ref{fig:app_output_template}. The \texttt{Solvability} and \texttt{Budget} fields are scalar values in $[0,1]$, representing the behavioral solvability value and the assigned budget fraction for the query. For \textit{nice fold} targets, \texttt{<predict>} carries a zero budget, the final answer is replaced with \texttt{<Unsolvable>}, and the \texttt{<think>} block is kept empty or reduced to a minimal abstention statement. The same template is retained throughout Stage~2 training and evaluation for \BET.

\begin{figure}[h]
\centering
\begin{betexample}[unbreakable]{Structured Output Template}
\begin{lstlisting}[style=betcode]
<predict>
Solvability: [0,1]
Budget: [0,1]
</predict>
<think>
... reasoning trace ...
</think>
\boxed{final answer}
\end{lstlisting}
\end{betexample}
\caption{Structured output template used in Stage~1 demonstrations and retained in Stage~2.}
\label{fig:app_output_template}
\end{figure}

\subsection{Example Stage~1 Targets}
\label{app:sft_example}

We illustrate how a raw query is converted into a Stage~1 target.

\begin{betexample}{Raw Query}
\begin{lstlisting}[style=betcode]
How many distinct arrangements of the letters in MISSISSIPPI have no two S's adjacent?
\end{lstlisting}
\end{betexample}

\vspace{4pt}

\begin{betexample}{Constructed \textit{Short Solve} Target}
\begin{lstlisting}[style=betcode]
<predict>
Solvability: 0.69
Budget: 0.24
</predict>
<think>
We want arrangements of the letters in MISSISSIPPI with no two S's adjacent.

First count the letters other than S. Removing the four S's leaves
M, I, I, I, I, P, P. These 7 letters can be arranged in

7! / (4! 2!)

ways, since the four I's are identical and the two P's are identical.

Now place the four S's into the gaps around this arrangement.
For any fixed arrangement of the non-S letters, there are 8 available
slots where an S can go: one before the first letter, one after the
last letter, and one between each pair of consecutive letters.

To ensure that no two S's are adjacent, we may place at most one S in
each slot. So we only need to choose 4 of the 8 slots for the 4
identical S's. The number of valid choices is

C(8,4).

Therefore the total number of valid arrangements is

7! / (4! 2!) * C(8,4).

Compute each factor:
7! / (4! 2!) = 5040 / 48 = 105,
C(8,4) = 70.

Hence the total number of arrangements is

105 * 70 = 7350.
</think>
\boxed{7350}
\end{lstlisting}
\end{betexample}

\vspace{4pt}

\textit{Hero call} targets use the same template with a larger budget value and a longer retained reasoning trace. \textit{Nice fold} targets instead take the following form.

\begin{betexample}{Constructed Nice Fold Target}
\begin{lstlisting}[style=betcode]
<predict>
Solvability: 0.00
Budget: 0.00
</predict>
<think>
This query is beyond my current reliable capability.
</think>
\boxed{<Unsolvable>}
\end{lstlisting}
\end{betexample}

These examples illustrate the Stage~1 supervision targets for \textit{short solve} and \textit{nice fold}.

\section{Full \BET Algorithm}
\label{app:algorithm}

For compactness, we introduce the following notation. Given a query $x$, let
$\textproc{Sample}(x;\pi)$ denote a single sample from policy $\pi$ together with its correctness flag, used
only for the Stage~1 cold start, and let $\textproc{Profile}(x;\pi,K,p)$ denote the online profiling operator of
Stage~2, which samples $K$ rollouts from $\pi$, computes the solvability estimate $\hat{s}_x$, the correct-rollout
set $\mathcal{C}_x$, and the efficient solution cost $\hat{c}^{*}_x$ from the shortest $p$-fraction of
$\mathcal{C}_x$. Let $\tilde{s}_x$ denote the behavioral solvability value written into a solvable Stage~1 target, obtained from
a monotone mapping of the query's difficulty label and the cost of its sampled solution.
Let $\textproc{Declare}(x;\pi,K)$ return the abstain-or-attempt decision of each of the $K$ rollouts, which \BET
commits in the \texttt{<predict>} block, with a zero budget marking abstention, before any reasoning is generated. Let $\textproc{BudgetTarget}(\hat{c}^{*}_x)$ denote the continuous budget target used for solvability calibration, and let
$\textproc{Format}(\hat{s},b,z,a)$ denote the structured output template consisting of a predicted
solvability $\hat{s}$, a predicted budget $b$, a reasoning trace $z$, and a final answer $a$.

Algorithm~\ref{alg:bet_stage1} gives the complete Stage~1 cold-start construction procedure. Every
profiled query is assigned. Queries the single sample fails go to \textit{nice fold},
and the remaining queries go to \textit{short solve} or \textit{hero call} according to
the cost of that sample before optional balancing. Algorithm~\ref{alg:bet_stage2} gives the full
Stage~2 GRPO training loop with dynamic solvability estimation and reward aggregation.

\begin{algorithm}[h]
\small
\caption{\BET Stage 1: Cold-Start Demonstration Construction}
\label{alg:bet_stage1}
\begin{algorithmic}[1]
\Require stratified pool $\mathcal{D}_{\mathrm{pool}}$, base policy $\pi_{\theta_0}$, split fraction $\rho$
\Ensure cold-start dataset $\mathcal{D}_{\mathrm{cs}}$

\State $\mathcal{S}_{\mathrm{solv}} \gets \varnothing$, \quad $\mathcal{S}_{\mathrm{unsolv}} \gets \varnothing$
\ForAll{$x \in \mathcal{D}_{\mathrm{pool}}$}
    \State $(y_x,\mathrm{ok}_x) \gets \textproc{Sample}(x;\pi_{\theta_0})$ \Comment{one sample per query}
    \If{$\neg\,\mathrm{ok}_x$}
        \State $\mathcal{S}_{\mathrm{unsolv}} \gets \mathcal{S}_{\mathrm{unsolv}} \cup \{x\}$ \Comment{zero-return regime}
    \Else
        \State $\mathcal{S}_{\mathrm{solv}} \gets \mathcal{S}_{\mathrm{solv}} \cup \{(x,y_x,c(y_x))\}$
    \EndIf
\EndFor

\State $q_{\mathrm{split}} \gets \textproc{Quantile}_{\rho}\bigl(\{c(y_x) : (x,y_x,c(y_x))\in\mathcal{S}_{\mathrm{solv}}\}\bigr)$
\State $\mathcal{D}_{\mathrm{cs}} \gets \varnothing$

\ForAll{$(x,y_x,c(y_x)) \in \mathcal{S}_{\mathrm{solv}}$}
    \State $b_x^{*} \gets \textproc{BudgetTarget}(c(y_x))$
    \If{$c(y_x) \le q_{\mathrm{split}}$}
        \State $t_x \gets \textproc{Format}(\tilde{s}_x, b_x^{*}, y_x, a_x^{\mathrm{box}})$
        \State $\mathcal{D}_{\mathrm{cs}} \gets \mathcal{D}_{\mathrm{cs}} \cup \{(x,t_x,\textit{Short Solve})\}$
    \Else
        \State $t_x \gets \textproc{Format}(\tilde{s}_x, b_x^{*}, y_x, a_x^{\mathrm{box}})$
        \State $\mathcal{D}_{\mathrm{cs}} \gets \mathcal{D}_{\mathrm{cs}} \cup \{(x,t_x,\textit{Hero Call})\}$
    \EndIf
\EndFor

\ForAll{$x \in \mathcal{S}_{\mathrm{unsolv}}$}
    \State $\hat{s}_x \gets 0$, \quad $b_x^{*} \gets 0$
    \State $t_x \gets \textproc{Format}(\hat{s}_x, b_x^{*}, \texttt{abstain}, \texttt{Unsolvable})$
    \State $\mathcal{D}_{\mathrm{cs}} \gets \mathcal{D}_{\mathrm{cs}} \cup \{(x,t_x,\textit{Nice Fold})\}$
\EndFor

\State $\mathcal{D}_{\mathrm{cs}} \gets \textproc{SubsampleOrBalance}(\mathcal{D}_{\mathrm{cs}})$ \Comment{optional class balancing}
\State \Return $\mathcal{D}_{\mathrm{cs}}$
\end{algorithmic}
\end{algorithm}

\begin{algorithm}[t]
\small
\caption{\BET Stage 2: GRPO with Dynamic Solvability Estimation}
\label{alg:bet_stage2}
\begin{algorithmic}[1]
\Require cold-start initialized policy $\pi_{\theta_1}$, training subset $\mathcal{D}_{\mathrm{sub}}$, rollout count $K$, percentile $p$, clipping coefficient $\epsilon_{\mathrm{clip}}$, normalization constant $\epsilon_{\mathrm{norm}}$, learning rate $\eta_{\mathrm{lr}}$, reward hyperparameters $(\delta,\lambda,\beta,\tau,\gamma_s,\gamma_b,\gamma'_s,\gamma'_b)$, abstention threshold $\epsilon_{\mathrm{abs}}$
\Ensure trained BET policy $\pi_{\theta^{*}}$

\State $\pi_{\theta} \gets \pi_{\theta_1}$
\While{not converged}
    \State $\pi_{\theta_{\mathrm{old}}} \gets \pi_{\theta}$ \Comment{rollout policy for this update}
    \State sample prompt batch $\mathcal{B} = \{x^{(1)},\ldots,x^{(B)}\} \sim \mathcal{D}_{\mathrm{sub}}$
    \ForAll{$x \in \mathcal{B}$}
        \State $\mathcal{A}_x \gets \textproc{Declare}(x;\pi_{\theta_{\mathrm{old}}},K)$ \Comment{abstain or attempt, read off the \texttt{<predict>} block}
        \If{every entry of $\mathcal{A}_x$ abstains}
            \State override one entry of $\mathcal{A}_x$, chosen uniformly at random, to attempt \Comment{guaranteed attempt}
        \EndIf
        \State $(\mathcal{Y}_x,\hat{s}_x,\mathcal{C}_x,\hat{c}^{*}_x) \gets \textproc{Profile}(x;\pi_{\theta_{\mathrm{old}}},K,p \mid \mathcal{A}_x)$
        \For{$k = 1$ to $K$}
            \State $y_k \gets \mathcal{Y}_x[k]$
            \State $\textproc{Parse}(y_k) \Rightarrow (\hat{\hat{s}}_k,\hat{b}_k,z_k,a_k)$
            \State $R_k^{\mathrm{VAL}} \gets R_{\textsc{val}}(y_k \mid x;\hat{s}_x,\epsilon_{\mathrm{abs}},\delta,\lambda)$
            \Comment{Eq.~\ref{eq:rval}}
            \State $R_k^{\mathrm{EFF}} \gets R_{\textsc{eff}}(y_k \mid x;\hat{s}_x,\hat{c}^{*}_x,\beta,\tau)$ \Comment{Eq.~\ref{eq:reff}}
            \State $R_k^{\mathrm{CAL}} \gets R_{\textsc{cal}}(y_k \mid x;\hat{s}_x,\hat{c}^{*}_x,\hat{\hat{s}}_k,\hat{b}_k)$ \Comment{Eq.~\ref{eq:rcal}}
            \State $\mathcal{R}_k(x) \gets R_k^{\mathrm{VAL}} + R_k^{\mathrm{EFF}} + R_k^{\mathrm{CAL}}$
        \EndFor

        \State $\bar{\mathcal{R}}_x \gets \frac{1}{K}\sum_{k=1}^{K}\mathcal{R}_k(x)$
        \State $\sigma_x \gets \mathrm{Std}(\{\mathcal{R}_k(x)\}_{k=1}^{K})$
        \For{$k = 1$ to $K$}
            \State $\hat{A}_k(x) \gets \dfrac{\mathcal{R}_k(x)-\bar{\mathcal{R}}_x}{\sigma_x+\epsilon_{\mathrm{norm}}}$ \Comment{group-relative normalization}
        \EndFor
    \EndFor

    \State form token-level clipped GRPO objective
    \[
    \mathcal{J}_{\mathrm{BET}}(\theta)
    \gets
    \mathbb{E}_{x \in \mathcal{B}}
    \left[
    \frac{1}{K}\sum_{k=1}^{K}
    \frac{1}{T_k}\sum_{t=1}^{T_k}
    \min\!\Big(
    r_{k,t}(\theta)\hat{A}_k(x),\;
    \mathrm{clip}(r_{k,t}(\theta),1-\epsilon_{\mathrm{clip}},1+\epsilon_{\mathrm{clip}})\hat{A}_k(x)
    \Big)
    \right],
    \]
    where $T_k$ is the number of generated tokens in $y_k$ and
    \[
    r_{k,t}(\theta)=
    \frac{\pi_{\theta}(y_{k,t}\mid x,y_{k,<t})}
         {\pi_{\theta_{\mathrm{old}}}(y_{k,t}\mid x,y_{k,<t})}.
    \]
    \State $\theta \gets \theta + \eta_{\mathrm{lr}} \nabla_{\theta}\mathcal{J}_{\mathrm{BET}}(\theta)$
\EndWhile

\State \Return $\pi_{\theta^{*}} \gets \pi_{\theta}$
\end{algorithmic}
\end{algorithm}

\paragraph{Guaranteed attempt within a group.}
The asymmetric gate of Eq.~\ref{eq:rval} takes effect only when a group contains an attempt. If all $K$ rollouts of a query abstain, then $\hat{s}(x)=0$ holds by construction and every rollout receives $+\delta$, so the group yields no reward contrast and the query stays in the zero set for as long as the policy keeps abstaining on it. We therefore constrain each group to retain one attempt. When all $K$ rollouts declare abstention, one of them, chosen uniformly at random, is overridden at the \texttt{<predict>} step before any reasoning is generated and completes a full reasoning trace. The group size stays at $K$, and the overridden decision is excluded from the policy-gradient term while its reasoning and answer are learned normally. On a query that remains beyond the policy, the forced attempt fails, $\hat{s}(x)$ stays at zero, and the abstaining rollouts keep their $+\delta$ against a penalized attempt, which sharpens the fold contrast. On a query that has become solvable, a single success lifts $\hat{s}(x)$ to $1/K$ and flips all abstentions to $-\lambda$.
 
Here $a_x^{\mathrm{box}}$ denotes the final answer in boxed form. Algorithm~\ref{alg:bet_stage1} corresponds to the cold-start construction of \S~\ref{sec:training} and Appendix~\ref{app:sft}, and Algorithm~\ref{alg:bet_stage2} to the Stage~2 loop of \S\S~\ref{sec:reward} and~\ref{sec:training}.

\section{Reward Ablation and Hyperparameter Sensitivity}
\label{app:ablation}

We further analyze each reward component in \BET and examine robustness to key hyperparameters. Our ablations cover four reward components: the solvability and budget calibration branches of $R_{\textsc{cal}}$, the abstention gate of $R_{\textsc{val}}$, and the query-adaptive efficiency reward $R_{\textsc{eff}}$. Within $R_{\textsc{val}}$ we ablate the fold component, since removing correctness as well collapses RL training. All ablations are conducted on Omni-Math~\citep{Gao2024OmniMATHAU} and MATH500~\citep{lightman2023lets} using Qwen3-4B-Thinking.

\subsection{Ablation of Reward Components}
\label{app:ablation_full}
Table~\ref{tab:app_ablation} reports the effect of removing each branch of $R_{\textsc{cal}}$ individually or jointly, the fold gate of $R_{\textsc{val}}$, and the efficiency reward $R_{\textsc{eff}}$.
Full \BET yields the best Pareto position on both benchmarks, while each ablation exposes a distinct failure mode.
\begin{table*}[t]
\centering
\small
\setlength{\tabcolsep}{4.6pt}
\renewcommand{\arraystretch}{1.12}
\caption{Ablation of the reward components on Omni-Math and MATH500, averaged over the same five stochastic evaluation runs as Table~\ref{tab:main_math}.
$R_{\textsc{cal}}^{\textsc{solv}}$ and $R_{\textsc{cal}}^{\textsc{bud}}$ denote the solvability and budget calibration branches of $R_{\textsc{cal}}$, $R_{\textsc{val}}^{\textsc{fold}}$ denotes the abstention gate of $R_{\textsc{val}}$, and $R_{\textsc{eff}}$ is the query-adaptive efficiency reward.}
\label{tab:app_ablation}
\begin{tabular}{l|cc|cc}
\toprule
\multirow{2}{*}{\textbf{Variant}}
& \multicolumn{2}{c|}{\textbf{Omni-Math}}
& \multicolumn{2}{c}{\textbf{MATH500}} \\
\cmidrule(lr){2-3}\cmidrule(lr){4-5}
& \textbf{Acc.\ (\%)} $\uparrow$ & \textbf{\#Tok.\ (K)} $\downarrow$
& \textbf{Acc.\ (\%)} $\uparrow$ & \textbf{\#Tok.\ (K)} $\downarrow$ \\
\midrule
Vanilla                                & 54.5 & 8.6  & 88.8 & 6.8 \\
w/o $R_{\textsc{cal}}^{\textsc{solv}}$ & 53.8 & 7.0  & 89.4 & 5.6 \\
w/o $R_{\textsc{cal}}^{\textsc{bud}}$  & 56.3 & 12.1 & 91.3 & 8.0 \\
w/o $R_{\textsc{cal}}$                 & 52.8 & 7.1  & 88.3 & 5.9 \\
w/o $R_{\textsc{val}}^{\textsc{fold}}$ & 53.4 & 7.5  & 88.8 & 6.2 \\
w/o $R_{\textsc{eff}}$                 & 55.8 & 6.2 & 90.2 & 5.3 \\
\textbf{Full \BET}                     & 55.6 & 3.8 & 90.0 & 3.3 \\
\bottomrule
\end{tabular}
\end{table*}

\paragraph{Removing the solvability branch $R_{\textsc{cal}}^{\textsc{solv}}$.}
This variant weakens \BET from a capability-aware allocator into a milder cost-sensitive compressor.
On Omni-Math, tokens drop from the vanilla 8.6K to 7.0K, a 19\% reduction, but the model no longer learns a reliable regime partition and accuracy slips below vanilla. Full \BET removes 56\% of the tokens while improving accuracy.
The same pattern appears on MATH500, where the model still reacts to cost but no longer aligns compression with current-policy solvability.

\paragraph{Removing the budget branch $R_{\textsc{cal}}^{\textsc{bud}}$.}
This produces the opposite failure mode.
The model retains a solvability signal and therefore continues to pursue hard-but-promising queries, but it loses the constraint that ties declared budget to empirical competence.
As a result, it over-requests compute almost everywhere.
This raises raw accuracy slightly, but at a prohibitive token cost. On Omni-Math, accuracy increases to 56.3\% while tokens surge to 12.1K, 41\% above the vanilla budget, and on MATH500 the same variant reaches 91.3\% accuracy but consumes 8.0K tokens, 17\% above vanilla.

\begin{figure}[H]
    \centering
    \includegraphics[width=0.85\linewidth]{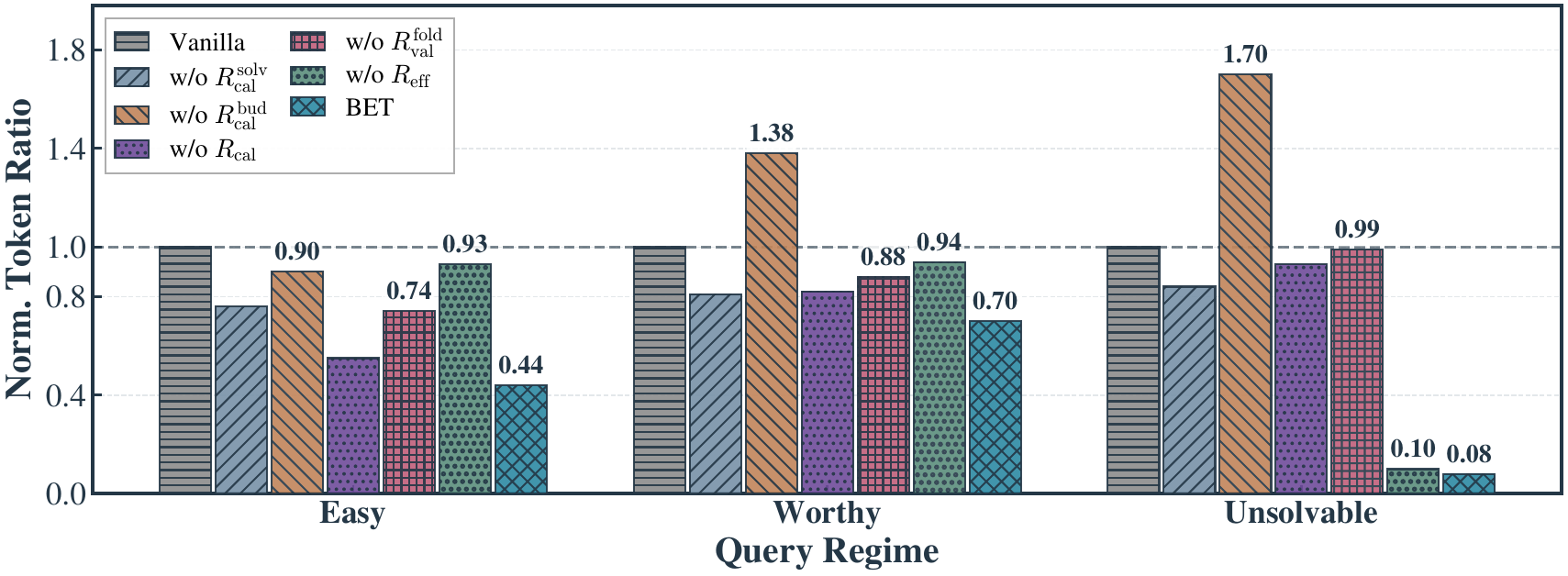}
\caption{Regime-wise token ratio on Omni-Math under reward-component ablations. The unsolvable cut requires the fold gate and calibration together. Only the two variants that keep both, full \BET and w/o $R_{\textsc{eff}}$, collapse the unsolvable regime to $0.08$ and $0.10$ of the vanilla budget, whereas removing the gate leaves $0.99$ and removing either calibration branch leaves $0.84$ to $1.70$. $R_{\textsc{eff}}$ controls compression on easy and worthy queries. Only full \BET separates all three regimes.}
    \label{fig:app_ablation_regime}
\end{figure}

\paragraph{Removing $R_{\textsc{cal}}$ entirely.}
When both calibration branches are removed, the model loses capability-aware partitioning and budget calibration, degrading into a coarse compressor driven only by $R_{\textsc{val}}$ and $R_{\textsc{eff}}$.
On Omni-Math, accuracy falls to 52.8\%, below vanilla, while tokens remain at 7.1K, only 17\% below the vanilla budget.
MATH500 follows the same pattern, with accuracy at 88.3\%, below vanilla, with tokens at 5.9K, a 13\% reduction.
Unlike single-branch ablations, which retain partial regime awareness, this variant shortens reasoning without separating the regimes (Fig.~\ref{fig:app_ablation_regime}), keeping unsolvable queries at $0.93$ of the vanilla budget against $0.82$ for worthy ones.

\paragraph{Removing the fold gate of $R_{\textsc{val}}$.}
The fold branch of $R_{\textsc{val}}$ grants $+\delta$ for abstaining on queries with $\hat s(x)<\epsilon_{\textsc{abs}}$ but penalizes premature surrender by $-\lambda$ with $\lambda \gg \delta$.
Removing this gate (i.e., setting $R_{\textsc{abstain}}\equiv 0$) eliminates the positive incentive for fold.
Within a few optimization steps, the policy stops emitting \texttt{<Unsolvable>} altogether and reverts to forced answering on every query, including those with $\hat s_0(x)=0$.
It is \emph{less} willing to fold than w/o $R_{\textsc{cal}}$, which still retains the $+\delta$ signal.
The effect is a modest accuracy increase from sporadic lucky guesses on previously folded queries (53.4\% on Omni-Math, 88.8\% on MATH500), paid for by a substantial token rebound on unsolvable cases (7.5K and 6.2K tokens, with the unsolvable regime ratio in Fig.~\ref{fig:app_ablation_regime} climbing back to $\approx\!1.0$), leaving 88\% and 91\% of the vanilla budget in place.
The fold gate is thus necessary for converting zero-return queries into near-zero token expenditure, and removing it reproduces the vanilla failure mode of treating unsolvable and worthy queries as similarly demanding. The gate alone does not produce the full profile. Fig.~\ref{fig:app_ablation_regime} shows that the variants which keep the gate but drop a calibration branch still spend 0.84 to 1.70 of the vanilla budget on unsolvable queries, so the cut also depends on $R_{\textsc{cal}}$ supplying a reliable solvability estimate.

\paragraph{Removing the efficiency reward $R_{\textsc{eff}}$.}
$R_{\textsc{eff}}$ shapes correct-trajectory length toward the efficient solution cost $\hat c^*(x)$.
Removing it ($\beta=0$) leaves $R_{\textsc{val}}$ and $R_{\textsc{cal}}$ intact, so fold and hero call behaviors persist. The model still abstains on zero-return queries and still calibrates declared budget against empirical cost.
What disappears is the pressure that compresses overthinking on easy and worthy queries, evidenced in Fig.~\ref{fig:app_ablation_regime} by the easy and worthy regime ratios climbing back to near-vanilla levels while the unsolvable ratio remains close to full \BET.
Consequently, accuracy is preserved or marginally improved (55.8\% on Omni-Math, 90.2\% on MATH500), since the policy is free to spend additional reasoning on solvable queries, but token consumption climbs to 6.2K and 5.3K. Both numbers stay well below vanilla because fold still suppresses the unsolvable regime, yet remain far above full \BET.
At 73\% and 78\% of the vanilla budget against 45\% and 48\% for full \BET, $R_{\textsc{eff}}$ is confirmed as the source of the query-adaptive compression that complements fold-driven waste reduction. Removing it leaves regime awareness intact but eliminates \textit{short solve} on easy queries.

\paragraph{Putting the four components together.}
Full \BET restores the strongest trade-off on both benchmarks, confirming that the four reward components play complementary roles. $R_{\textsc{val}}^{\textsc{fold}}$ activates abstention on zero-return queries, $R_{\textsc{eff}}$ compresses redundant reasoning on solvable ones, $R_{\textsc{cal}}^{\textsc{solv}}$ supplies the solvability signal needed for regime separation, and $R_{\textsc{cal}}^{\textsc{bud}}$ anchors planned compute to the empirical cost profile.
The two new ablations form a clean dual. Removing $R_{\textsc{val}}^{\textsc{fold}}$ collapses the unsolvable regime back toward vanilla while leaving easy and worthy regimes partially compressed, whereas removing $R_{\textsc{eff}}$ preserves the unsolvable cut but allows easy and worthy queries to drift back toward overthinking.
The three behaviors of~\S\ref{sec:emergence} depend on different subsets of the reward. \textit{Short solve} requires efficiency shaping on solvable queries, \textit{nice fold} requires an explicit abstention value on zero-return queries, and \textit{hero call} requires calibration signals that prevent solvable hard queries from being folded or over-compressed. Reproducing the sharp allocation profile reported in~\S\ref{sec:emergence} requires all four components acting jointly.

\subsection{Sensitivity to Key Hyperparameters}
\label{app:sensitivity}
We further examine the sensitivity of \BET to three key hyperparameters: the fold reward $\delta$, the efficiency bonus strength $\beta$, and the efficient-cost percentile $p$. Table~\ref{tab:app_sensitivity} reports the resulting trade-offs. Increasing $\delta$ encourages more aggressive folding, which trades a small amount of accuracy for a lower token budget. Increasing $\beta$ strengthens compression on solvable queries, improving efficiency at the cost of a mild accuracy trade-off. Moving $p$ away from its default costs in either direction. At $20\%$ the cost estimate is too tight and accuracy falls by up to $2.2$ points, while at $40\%$ it is too permissive and tokens rise by a tenth to a fifth with no accuracy gain. Overall, $\delta$ and $\beta$ trade accuracy against budget smoothly, $p$ has a clear optimum at the default, and that setting gives the best cost-quality balance on all three benchmarks.

\begin{table*}[t]
\centering
\small
\setlength{\tabcolsep}{4.0pt}
\renewcommand{\arraystretch}{1.10}
\caption{Sensitivity of \BET to key hyperparameters on Omni-Math, MATH500, and AIME-25. Unless varied, all hyperparameters are fixed to the default setting $(\delta,\beta,p)=(0.10,0.30,30\%)$. All entries are averaged over the same five stochastic evaluation runs as Table~\ref{tab:main_math}.}
\label{tab:app_sensitivity}
\begin{tabular}{cc cc cc cc}
\toprule
\multirow{2}{*}{\textbf{Param.}} & \multirow{2}{*}{\textbf{Value}}
& \multicolumn{2}{c}{\textbf{Omni-Math}}
& \multicolumn{2}{c}{\textbf{MATH500}}
& \multicolumn{2}{c}{\textbf{AIME-25}} \\
\cmidrule(lr){3-4}\cmidrule(lr){5-6}\cmidrule(lr){7-8}
& & \textbf{Acc.\ (\%)} $\uparrow$ & \textbf{\#Tok.\ (K)} $\downarrow$
  & \textbf{Acc.\ (\%)} $\uparrow$ & \textbf{\#Tok.\ (K)} $\downarrow$
  & \textbf{Acc.\ (\%)} $\uparrow$ & \textbf{\#Tok.\ (K)} $\downarrow$ \\
\midrule
\multirow{3}{*}{$\delta$}
& 0.02          & 56.2 & 6.1 & 90.8 & 4.7 & 66.7 & 9.4 \\
& 0.05          & 55.9 & 5.0 & 90.8 & 4.2 & 66.0 & 7.7 \\
& \textbf{0.10} & \textbf{55.6} & \textbf{3.8} & \textbf{90.0} & \textbf{3.3} & \textbf{64.7} & \textbf{6.1} \\
\addlinespace[2pt]
\multirow{3}{*}{$\beta$}
& 0.20          & 55.7 & 4.2 & 90.1 & 3.6 & 64.7 & 6.8 \\
& \textbf{0.30} & \textbf{55.6} & \textbf{3.8} & \textbf{90.0} & \textbf{3.3} & \textbf{64.7} & \textbf{6.1} \\
& 0.40          & 55.1 & 3.6 & 89.8 & 3.1 & 62.7 & 5.7 \\
\addlinespace[2pt]
\multirow{3}{*}{$p$}
& 20\%          & 53.4 & 3.7 & 87.9 & 3.2 & 63.3 & 6.0 \\
& \textbf{30\%} & \textbf{55.6} & \textbf{3.8} & \textbf{90.0} & \textbf{3.3} & \textbf{64.7} & \textbf{6.1} \\
& 40\%          & 55.5 & 4.2 & 89.9 & 3.8 & 62.7 & 7.2 \\
\bottomrule
\end{tabular}
\end{table*}

\section{Fold Decisions and Preserved Exploration}
\label{app:fold_discussion}

\subsection{Design Goal}

\BET chooses among solving a query concisely, exploring it in depth, and declining to answer, so that stopping an unproductive investment is a behavior the policy selects for itself.
A fold reports low expected return under the current model, budget, and reward configuration, and a stronger model or a later checkpoint may still solve the query.
A query is a solving opportunity whenever a correct path exists that the current policy could find, and it keeps that status whether the path is long or seldom sampled.
Finite sampling that misses such a path produces an error of identification, and \S~\ref{app:fold_sampling} measures how large it is.
The design goal is to reduce futile attempts while preserving as much as possible of the successful reasoning the current policy can realize, including solutions that are expensive to reach or rarely found.

\subsection{How the Reward Preserves Exploration}

Four properties follow from that goal.
Abstention is gated asymmetrically, worth $+\delta$ in a group with no success and $-\lambda$ with $\lambda \gg \delta$.
The evidence required is one rollout, since $\epsilon_{\mathrm{abs}} = 1/K$ is the lowest threshold the estimator can express, so nothing in the design asks a query to be easy before it is worth attempting.
Correct solutions receive $+1$ at any length, and $R_{\textsc{eff}}$ rewards those under $\hat{c}^{*}(x)$ without penalizing those above it, so depth on a hard query is not itself costly.
Finally, when all $K$ rollouts abstain one is overridden at the \texttt{<predict>} step (Appendix~\ref{app:algorithm}), so a success stays observable.

These settings also bound how low solvability may fall before the reward stops preferring an attempt.
Measured against a fixed abstention value of $\delta$, attempting a query with true solvability $s$ has expected value $s - (1-s)\,\alpha_{\mathrm{fail}}\hat{c}$ under Eq.~\ref{eq:rval}, with $\hat{c}=c/L_{\max}$ the share of the completion budget consumed, so an attempt is preferred when
\begin{equation}
\label{eq:breakeven}
s \;>\; s^{*}(\hat{c}) \;=\; \frac{\delta + \alpha_{\mathrm{fail}}\hat{c}}{1 + \alpha_{\mathrm{fail}}\hat{c}}.
\end{equation}
With $\delta = 0.10$ and $\alpha_{\mathrm{fail}} = 0.20$ this reference threshold is $0.14$ at a quarter of the budget and $0.25$ at the full budget.
The gate makes the actual comparison more favorable to attempting still, because abstention is worth $+\delta$ only when no rollout in the group succeeds. Against $n$ attempting peers of success rate $q$ its expected value is $(\delta+\lambda)(1-q)^{n}-\lambda$, which at $n{=}15$ and $q{=}0.2$ is $-0.77$ against $+0.04$ for a full-budget attempt.
Eq.~\ref{eq:breakeven} therefore bounds from above the solvability at which the reward stops preferring an attempt, and $\delta$ moves that bound, which is why it acts as a single accuracy-efficiency knob (\S~\ref{sec:main_results}, Appendix~\ref{app:sensitivity}).

Table~\ref{tab:app_control} places the design among existing ways of controlling test-time compute.
Among the baselines, only KWTQ abstains, declining below a threshold on an auxiliary value head as a decoding-time rule outside the training objective.
In \BET the decision is taken inside the policy and graded by the reward, with the value of abstaining set by whether success evidence appears and trained jointly with the cost estimate and the budget calibration.

\begin{table}[t]
\centering
\small
\setlength{\tabcolsep}{5pt}
\renewcommand{\arraystretch}{1.15}
\caption{\textbf{How compute-control mechanisms allocate and stop reasoning.} \textit{Reasoner training}: the reasoning model's own parameters are updated. \textit{Length-aware training}: the objective carries a generation-cost term. \textit{Explicit adaptive control}: allocation is conditioned on the individual query. \textit{Abstain action}: declining to answer is a possible outcome. Row colors mark the four families of Tab.~\ref{tab:main_math}.}
\label{tab:app_control}
\resizebox{\textwidth}{!}{%
\begin{tabular}{@{}ll cccc@{}}
\toprule
\textbf{Method} & \textbf{Controlling signal} & \makecell{Reasoner\\training} & \makecell{Length-aware\\training} & \makecell{Explicit adaptive\\control} & \makecell{Abstain\\action} \\
\midrule
\rowcolor{c1bg} DiffAdapt & Difficulty predicted before decoding & \no & \no & \yes & \no \\
\rowcolor{c2bg} DEER & Decoding confidence & \no & \no & \yes & \no \\
\rowcolor{c2bg} KWTQ & Auxiliary value head & \no & \no & \yes & \yes \\
\rowcolor{c3bg} VeriThinker & Static verification supervision & \yes & \no & \no & \no \\
\rowcolor{c3bg} Overthink & Static short-chain supervision & \yes & \yes & \no & \no \\
\rowcolor{c4bg} Length-Penalty & Uniform length cost & \yes & \yes & \no & \no \\
\rowcolor{c4bg} CODA & Online difficulty estimate & \yes & \yes & \yes & \no \\
\rowcolor{c4bg} DR.SAF & Grouped pass rate & \yes & \yes & \yes & \no \\
\rowcolor{betbg} \textbf{\BET}~\textit{(ours)} & Group success $\hat{s}(x)$ and cost $\hat{c}^{*}(x)$ & \yes & \yes & \yes & \yes \\
\bottomrule
\end{tabular}}
\end{table}

\subsection{Evidence from the Experiments}

A fold counts as incorrect throughout our evaluation and receives no accuracy credit.
Under the vanilla-derived partition of \S~\ref{sec:regime}, \BET spends $0.08$ of the vanilla budget on queries vanilla never solves, against $0.76$ to $0.99$ across the baselines (Fig.~\ref{fig:fig4}(a)), so the savings of \S~\ref{sec:main_results} concentrate in the regime the fold action targets.
The worthy regime retains $0.70$ of the budget, and there \BET makes the largest net correctness gain, adding $48$ solved instances where Length-Penalty and DR.SAF lose $47$ and $3$. It also recovers $5$ on queries vanilla never solved (Fig.~\ref{fig:fig4}(b)).
Accuracy rises under the reduced budget, consistent with folds landing mostly on queries that would not have been solved.

The ablations in Appendix~\ref{app:ablation} show that this allocation depends on the reward as a whole: removing $R_{\textsc{val}}^{\textsc{fold}}$ returns the unsolvable regime to $0.99$, removing $R_{\textsc{eff}}$ holds the cut at $0.10$ but lets easy and worthy queries drift back toward overthinking, and removing either calibration branch leaves the unsolvable regime between $0.84$ and $1.70$.
In the settings evaluated, \BET reaches a more favorable accuracy-cost trade-off by cutting computation where it does not pay, while retaining effective reasoning on queries that remain solvable.

\subsection{Decision Reliability under Finite Sampling}
\label{app:fold_sampling}

We first measure how stable $K$-rollout labels are.
Under nested rollout budgets on four benchmarks (Fig.~\ref{fig:fig5}(c)), $19.31\%$ of empirical-zero labels acquire a success between $K{=}8$ and $16$, $0.43\%$ between $16$ and $24$, and none between $24$ and $32$, so labels that survive a group of $16$ are largely stable against further sampling.
The rate follows from the estimator. An empirical zero requires all $K$ rollouts to fail, an event of probability $(1-s)^{K}$, which at $K{=}16$ falls below $5\%$ once $s \geq 0.17$ and below $0.3\%$ once $s \geq 0.30$.

What this leaves is the low-solvability tail.
When true solvability is very small an empirical zero remains a likely outcome, so a rare success can still go unobserved. Such queries also carry the lowest expected return under Eq.~\ref{eq:objective}, so the residual costs little.
$K$ governs whether success evidence can be observed and the asymmetric gate governs how abstention is treated once it has been. We set $K{=}16$ at the point where additional rollouts stop repaying their cost.

\section{Limitations}
\label{app:limitations}

\BET has several limitations. First, training relies on $K$-rollout grouped sampling for online solvability estimation, and this grouped rollout requirement may become costly for larger models or longer contexts. Second, the current solvability estimate treats each query independently and does not model cross-query structure or curriculum effects during training. Third, our formulation focuses on single-query reasoning with explicit token-budget trade-offs. Extending the same investment-based principle to multi-turn agent settings, tool use, or environments with delayed external feedback remains an important direction for future work.

\section{Broader Impacts}
\label{app:broader}

This work studies how to reduce unnecessary test-time computation in large reasoning models while preserving problem-solving capability. A positive impact is that more efficient reasoning can reduce inference cost, latency, and energy consumption, making reasoning models more accessible to users and organizations with limited computational budgets. The learned \textit{nice fold} behavior may also expose capability-aware uncertainty, which can be useful in cascaded systems where difficult queries are escalated to stronger models.

Lowering the cost of reasoning models can also make both beneficial and harmful uses easier to scale, including automated generation of misleading content, spam, or low-quality decision support. In addition, an incorrectly calibrated fold decision may withhold useful answers, while an overconfident answer after insufficient reasoning may mislead users. These risks are especially important in high-stakes domains such as medicine, law, finance, or safety-critical decision making. Deployment in sensitive settings should include domain-specific validation, monitoring, human oversight, and appropriate escalation policies.

\end{document}